\pdfoutput=1

\documentclass[11pt]{article}

\usepackage[preprint]{acl}

\usepackage{times}
\usepackage{latexsym}

\usepackage[T1]{fontenc}

\usepackage[utf8]{inputenc}

\usepackage{microtype}

\usepackage{inconsolata}

\usepackage{graphicx}

\usepackage[most]{tcolorbox}

\usepackage{booktabs}

\usepackage{comment}

\usepackage{url}

\usepackage{array}

\usepackage{amsfonts}
\usepackage{amsmath}

\usepackage{float}

\usepackage{multirow}

\usepackage{enumitem,kantlipsum} 

\title{Steering Language Models in Multi-Token Generation:\\A Case Study on Tense and Aspect}

\author{
\textbf{Alina Klerings\textsuperscript{1}},
\textbf{Jannik Brinkmann\textsuperscript{2}},
\textbf{Daniel Ruffinelli\textsuperscript{1}},
\textbf{Simone Paolo Ponzetto\textsuperscript{1}}
\\
\\
\textsuperscript{1}University of Mannheim,
\textsuperscript{2}Technical University Clausthal
\\
\small{\texttt{alina.klerings@uni-mannheim.de}} \\
}

\begin{document}
\maketitle
\begin{abstract}
Large language models (LLMs) are able to generate grammatically well-formed text, but how do they encode their syntactic knowledge internally? While prior work has focused largely on binary grammatical contrasts, in this work, we study the representation and control of two multidimensional hierarchical grammar phenomena—verb tense and aspect—and for each, identify distinct, orthogonal directions in residual space using linear discriminant analysis. Next, we demonstrate causal control over both grammatical features through concept steering across three generation tasks. Then, we use these identified features in a case study to investigate factors influencing effective steering in multi-token generation. We find that steering strength, location, and duration are crucial parameters for reducing undesirable side effects such as topic shift and degeneration. Our findings suggest that models encode tense and aspect in structurally organized, human-like ways, but effective control of such features during generation is sensitive to multiple factors and requires manual tuning or automated optimization.\footnote{\url{https://github.com/klerings/tense-aspect}}
\end{abstract}

\section{Introduction}
\label{sec:introduction}
Growing evidence on the generative capabilities of large language models (LLMs) suggests that they encode structural properties of language---such as syntax trees---within their hidden representations \citep{hewitt-manning-2019-structural, diego2024polar}. Studying the representation of these properties can reveal how syntactic structures in models compare to those in human language. It may also help develop more linguistically aware AI systems, extend model capabilities to low-resource languages, improve the interpretability of generation, and assist in diagnosing systematic errors in tasks like machine translation \citep{lopez2025linguistic}.

\begin{figure}[t!]
    \centering
    \vspace{-0.8cm}
    \includegraphics[width=1\linewidth]{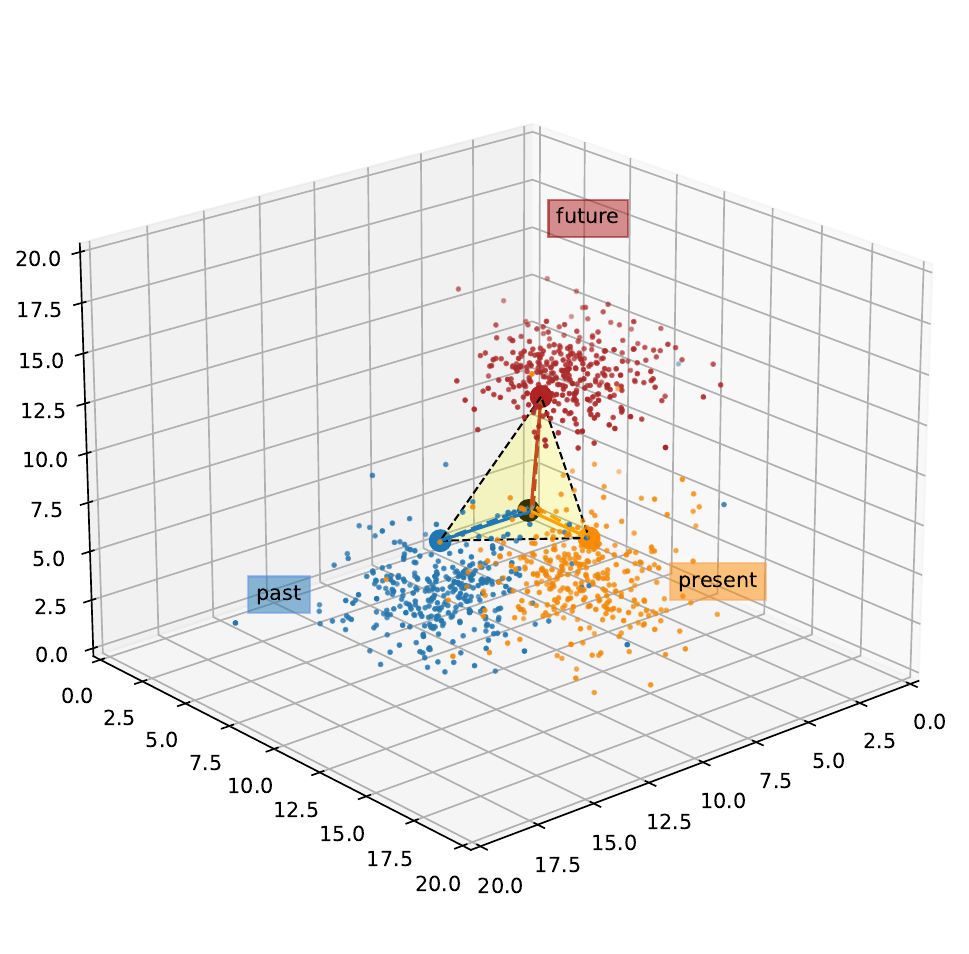}
    \vspace{-0.8cm}
    \caption{Activations from Qwen-7B projected along three identified feature directions that represent the categorical values of \textbf{tense}: present, past and future.}
    \label{fig:lda_tense_dirs}
\end{figure}

Prior work has assessed grammatical competence with behavioral evaluations, using black-box measures such as perplexity \citep{chen2024sudden} and grammatical acceptability benchmarks \citep{warstadt2020blimp, hu2020systematic}. However, these methods do not reveal how grammatical concepts are internally represented. To address this, recent interpretability studies have examined the mechanisms underlying syntactic competence, such as indirect object identification \citep{wang2023interpretability} and subject-verb agreement \citep{ferrando-costa-jussa-2024-similarity}, providing detailed insights into the role of specific architectural components. Yet, they largely focus on binary grammatical contrasts (e.g., singular vs. plural agreement), and often evaluate on single-token continuations, which limit their ability to capture more compositional or distributed grammatical phenomena. 

In this work, we take a complementary approach focused on grammatical \textbf{tense} and \textbf{aspect}, two core verb properties that encode temporal relations and often span multiple tokens. While tense positions an event in different time periods relative to the moment of speech (i.e., past, present or future), 

\begin{quote}
    a) She \textit{drove} her car.\newline
    b) She \textit{drives} her car.\newline
    c) She \textit{will drive} her car.
\end{quote}

aspect introduces an additional layer of temporal structure---describing whether an event is completed, ongoing or repeated (e.g., simple, perfect, progressive or perfect progressive) \citep{klein2009time}.

\begin{quote}
    d) She \textit{has driven} her car.\newline
    e) She \textit{is driving} her car.\newline
    f) She \textit{has been driving} her car.
\end{quote}

Unlike binary syntactic features, tense and aspect involve multiple discrete categories and can be combined (e.g., past perfect). Theoretically, they are independent grammatical features: any aspect should, in principle, be expressible in any tense. This separability is useful for a mechanistic analysis, as it allows for a systematic exploration of how these categories interact. We locate and analyze the representation of both properties in two models via linear probes and Linear Discriminant Analysis (LDA) \citep{park2024geometry}, and test their causal relevance \citep{vig2020investigating, mueller2024quest} by using them to steer LLM outputs in multi-token generation.

While prior work has demonstrated representational steering for single-token continuations and has focused on semantic properties, our approach applies steering to full sentences by intervening on every generated token individually, and focuses on grammatical phenomena. Since steering multiple tokens increases the risk for degenerate outputs and unintended side effects \citep{bricken2023monosemanticity, stickland2024steering}, we analyze when and how strongly to intervene. We find that steering effectiveness depends on activation norms, model architecture, and the nature of the task.

\paragraph{Contributions.} Our key contributions are as follows:

\begin{itemize}[leftmargin=*]
    \item We identify distinct, near-orthogonal directions in LLM residual space encoding tense and aspect, using probes and LDA.
    \item We demonstrate causal control over these grammatical features through concept steering, thereby mediating the LLM output during multi-token generation.
    \item We investigate conditions for effective steering, analyzing the influence of steering strength and location, activation norm, model type and task on side effects.
\end{itemize}

\section{Background and Related Work}
\label{sec:background}
\paragraph{Linear Representation Hypothesis and Concept Directions.}
The linear representation hypothesis is the assumption that abstract features are encoded as linear directions in LLM residual space \citep{elhage2022toy, nanda-etal-2023-emergent, park2023linear, marks2024the, park2024geometry}. This has led to the notion of \textbf{feature directions}—unit vectors in activation space corresponding to specific properties—and, more generally, \textbf{feature subspaces}, which span sets of related directions (e.g., different tense directions in shared tense subspace) \citep{geigercausalabstractions, mueller2024quest}.

Such representations have been extracted using sparse autoencoders \citep{bricken2023monosemanticity, huben2024sparse}, dimensionality reduction \citep{gurnee2024language}, \citep{heinzerling2024monotonic}, and supervised linear probes \citep{marks2024the}.  Recently, \citet{park2024geometry} introduced an LDA-based method to identify scaled directions that capture categorical structure. Unlike prior approaches focused on concepts with natural opposites (e.g. \textsc{male} vs. \textsc{female}), this framework models categorical concepts (e.g., \textsc{tense}) as sets of \textbf{binary features} (e.g., \textsc{future}, \textsc{present}, \textsc{past}), independently of predefined class structures.

\paragraph{Steering Language Models.} Feature steering has mainly targeted semantic features like sentiment \citep{rimsky-etal-2024-steering, lee2025programming} or numbers \citep{heinzerling2024monotonic}, while categorical linguistic properties remain underexplored.  Earlier work has focused on single-token interventions, with recent studies beginning to explore steering full-sentence generation \citep{lee2025programming, rimsky-etal-2024-steering, wu2025axbenchsteeringllmssimple}. Systematic evaluations, e.g., of steering strength, side effects, or degeneration, are still sparse \citep{pres2024reliable}.

Our work addresses these gaps by steering categorical grammatical features---specifically, tense and aspect---during sentence-level generation with controlled intervention strength and position-aware application. We extend the LDA-based framework from \citet{park2024geometry} by broadening its application from lexical to sentence-level grammatical categories. 

\paragraph{Grammatical Knowledge in LMs.} The grammatical competence of language models has been an ongoing research topic since before the emergence of LLMs, with subfields such as "Bertology" \citep{rogers-etal-2020-primer} focusing on earlier architectures like BERT. Tense has been explored in behavioral evaluations as well as causal analyses \citep{merullo2024language, zhang2025the}, but typically via binary distinctions (e.g., past vs. present) and single-token interventions---aside from \citet{brinkmann2025large}, who consider open-ended generation. However, tense and aspect have not been jointly analyzed in a unified framework. We address this by studying these features in multi-token generation, combining probing, representation space analysis, and causal steering to examine how grammatical concepts are encoded and can be controlled. For a more comprehensive review of related work, see App.~\ref{app:related_work}.

\section{Locating Tense and Aspect}
\label{sec:tense_aspect}
We focus on the interlinked grammatical concepts of verb tense and aspect by localizing and visualizing their subcategories via exploratory methods.

\subsection{Experimental Setup}
\label{sec:localization}
Tense and aspect are core grammatical categories that encode temporal structure. Tense refers to the time at which an event occurs (present, past, future), while aspect characterizes its temporal structure (simple, progressive, perfect, perfect progressive). In theory, each tense can occur in combination with each aspect, yielding a grid of 12 distinct tense-aspect forms (see App.~\ref{app:tense_aspect_overview}). This system makes these features ideal for studying multidimensional hierarchical structure in LLMs. In practice, such regular and compositional combinations are rare across languages. The comparatively structured tense-aspect system in English is an exception \citep{klein2009time}. Therefore, we focus our study on English rather than languages with less regularity.

\paragraph{Data.} We analyze sentences annotated with grammatical tense and aspect, requiring each sentence to contain a single, unambiguous tense-aspect combination to isolate a clear signal for each target variable. We use sentences from the Penn Treebank \citep{marcus-etal-1993-building} annotated with PropBank \citep{kingsbury-palmer-2002-treebank}, which provides verb-specific tense and aspect labels. 
After filtering out ambiguous sentences, the class distribution is highly skewed, with most examples in the simple aspect and some targets having fewer than ten instances. To address this imbalance, we augment rare classes with synthetic examples generated by GPT-4o \citep{hurst2024gpt}, resulting in 1,543 labeled sentences. Augmentation improves balance, but some categories remain overrepresented (e.g., simple past), therefore we downsample them for training the classifiers. For evaluation, we use the verb tense subset of 281 sentences from BIG-bench \citep{srivastava2023beyond, logeswaran2018content} (see App.~\ref{app:dataset} for dataset details).

\paragraph{Models.} We use the models Llama-3.1-8B-Instruct \citep{grattafiori2024llama3herdmodels}, primarily trained on English data, and Qwen-2.5-7B-Instruct \citep{qwen2025qwen25technicalreport}, with a focus on English and Chinese, that are commonly studied for similar analyses. For brevity, we omit version numbers and the ‘Instruct’ suffix throughout the remainder of the paper.

\paragraph{Localization via Probing.}  To localize the target concepts---tense, aspect and their combination---we train linear probing classifiers \citep{belinkov-2022-probing} on the hidden representations of the pre-trained LLMs. Let the model be defined as $f_{\theta}: x \rightarrow h$, where $\theta$ are learnable parameters, $x = (x_1, ..., x_N)$ the input tokens, and $h^l = (h_1^l, ..., h_N^l)$ the hidden representations at layer $l$. At each layer, we train a probing model $p^l: h^l_{\textrm{agg}} \rightarrow y$ that maps aggregated hidden states $h^l_{\textrm{agg}}$ to the corresponding tense, aspect or tense-aspect labels $y$ using multinomial logistic regression. We compute $h_{\textrm{agg}}$ per layer as follows, 
\begin{equation}
    h_{\textrm{agg}} = \frac{1}{\sqrt{N}}\sum_{i=1}^N h_i,
\end{equation} 
summing token-level representations and normalizing by the square root of sequence length $N$, a strategy that outperforms other aggregation methods, see App.~\ref{app:probing_heatmaps} for more details. Finally, we mean-center the aggregated hidden representations before classification.

\paragraph{Representation Geometry.}
To analyze how tense and aspect are represented in model activations, we use the framework proposed by \citet{park2024geometry} for handling categorical features. In this approach, each categorical concept (e.g., \textsc{tense}) consists of a set of subordinate feature values (e.g., \textsc{past}, \textsc{present}, \textsc{future}). To model each categorical value as a direction, they are represented as \textbf{binary features}: \{\textsc{is\_past}, \textsc{is\_not\_past}\}, \{\textsc{is\_present}, \textsc{is\_not\_present}\}, \{\textsc{is\_future}, \textsc{is\_not\_future}\}. 
Following this framework, we estimate vector representations for each feature value using the variant of LDA proposed by \citet{park2024geometry}. The objective is to find a vector that reduces within-class variation and highlights differences to all other classes.
Unlike standard LDA, which relies on both within-class and between-class covariance, this variant omits the latter, enabling the computation of each vector independently of the others. 

Formally, for each binary feature $w$, we compute a \textbf{normalized class direction} $\tilde{h}_w$ from the empirical mean of the class-specific activations $\mathbb{E}(h_w)$, adjusted by the pseudo-inverse of the class covariance $\mathrm{Cov}(h_w)^{\dag}$:

\begin{equation}
    \tilde{h}_w = \frac{\mathrm{Cov}(h_w)^{\dag}\mathbb{E}(h_w)}{\|\mathrm{Cov}(h_w)^{\dag}\mathbb{E}(h_w)\|_2}.
\end{equation}

This unit vector captures the direction of the class in residual space. To incorporate the strength of the signal, we scale the direction by the projection of the class mean onto it:

\begin{equation}
    \bar{\ell}_w = (\tilde{h}_w^\top\mathbb{E}(h_w))\tilde{h}_w.
\end{equation}

The resulting vector $\bar{\ell}_w$ encodes the orientation and intensity of the concept in activation space. 

Importantly, because each vector is computed independently, the method avoids enforcing any pre-defined class structure of tense and aspect. Instead, the representational geometry that emerges reflects the structure learned by the model itself.

Another key concept introduced by \citet{park2024geometry} is \textbf{binary contrast}, which captures the distinction between two categorical values within the same parent category and is computed as the vector difference between their feature vectors. In our analysis, we use binary contrasts to model categories in a lower-dimensional space. This allows us to (i) compare subordinate features within a category where their number exceeds the available representational dimensions (i.e., for aspectual values), and (ii) approximate latent dimensions for cross-category comparisons between tense and aspect.

\subsection{Results}

\paragraph{Representations of tense and aspect emerge early in the model.} Our probing classifiers predict tense and aspect from the embedding layer with f1-scores above 90\% and improve further with model depth, especially for the fine-grained tense-aspect combinations, see Table~\ref{tab:probing_acc}. 

\newcolumntype{P}[1]{>{\centering\arraybackslash}p{#1}}
\begin{table}[!h]
    \centering
    \small
    
    \setlength{\tabcolsep}{4pt} 
    \begin{tabular}{l|P{0.1\linewidth}P{0.13\linewidth}P{0.25\linewidth}}
     & \textbf{Tense} & \textbf{Aspect} & \textbf{Tense-Aspect} \\
     \midrule
        \small{Llama-8B} & 1.0 & 0.98 & 0.93 \\
        \small{Qwen-7B} & 1.0 & 0.98 &  0.92 \\
    \end{tabular}
    \caption{Target-wise F1-scores from the best-performing probe across layers.}
    \label{tab:probing_acc}
\end{table}

This demonstrates that contextualization is beneficial, as many tenses and aspects are expressed across multiple tokens. For detailed results across all layers and more aggregation strategies, see App.~\ref{app:probing_heatmaps}. The findings are consistent across targets and models and similar to earlier work that suggests syntactic processing happens before more complex semantic processing \citep{he-etal-2024-decoding}.

\paragraph{Grammatical properties form subspaces in representation space.} To analyze the LDA results, we use the 2D and 3D visualizations of \citet{park2024geometry}. To assess whether hidden representations encode a separation between categories of a single grammatical feature, we project test set embeddings from Qwen-7B onto selected directions.

For tense, we use $\bar{\ell}_{\text{\scriptsize\textsc{present}}}$, $\bar{\ell}_{\text{\scriptsize\textsc{past}}}$ and $\bar{\ell}_{\text{\scriptsize\textsc{future}}}$ as projection axes. For aspect, we use binary contrasts retrieved from vector differences to represent the four categories: $\bar{\ell}_{\text{\scriptsize\textsc{progressive}}}-\bar{\ell}_{\text{\scriptsize\textsc{simple}}}$, $\bar{\ell}_{\text{\scriptsize\textsc{perfect}}}-\bar{\ell}_{\text{\scriptsize\textsc{simple}}}$ and $\bar{\ell}_{\text{\scriptsize\textsc{perfect progressive}}}-\bar{\ell}_{\text{\scriptsize\textsc{simple}}}$.

\begin{figure}[H]
    \centering
    \includegraphics[width=1\linewidth]{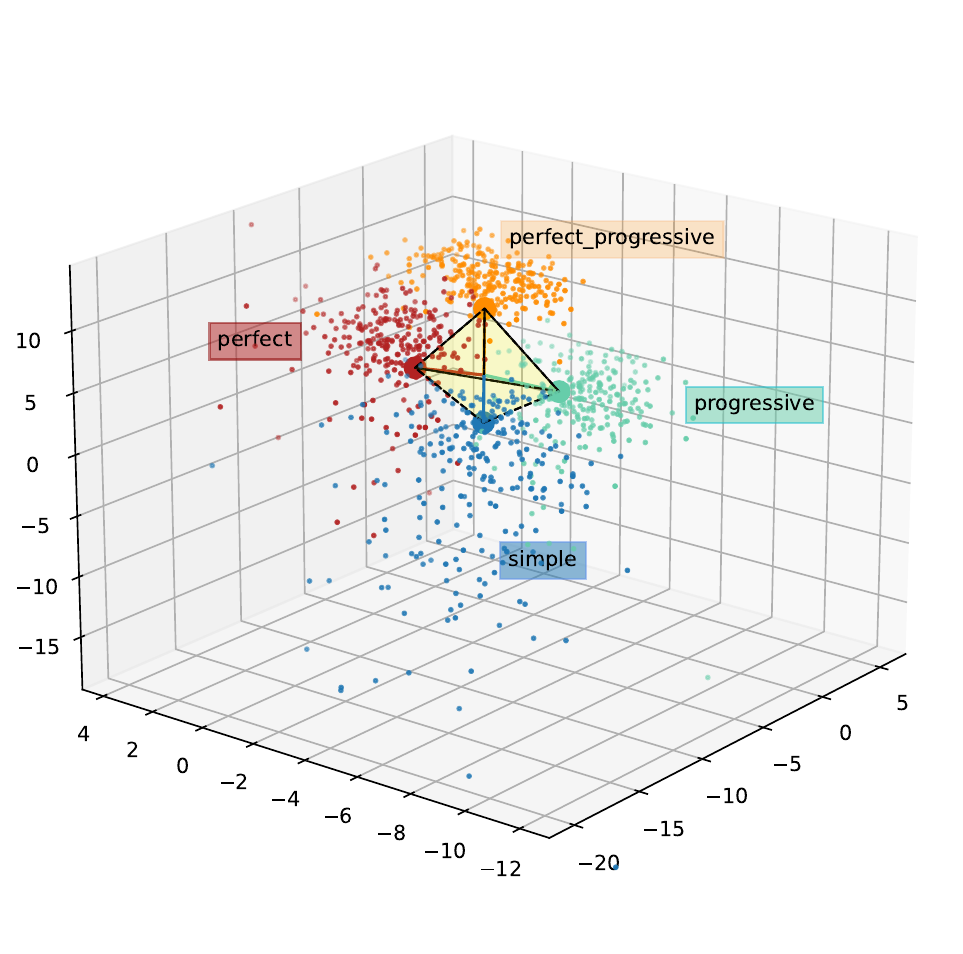}
    \vspace{-0.8cm}
    \caption{Projection of Qwen-7B hidden states (L0) along LDA-based \textbf{aspect} directions.}
    \label{fig:lda_aspect_dirs}
\end{figure}

In both cases, 3D projections reveal distinct clusters corresponding to the underlying grammatical categories: three well-separated clusters for tense (Figure~\ref{fig:lda_tense_dirs}, explained variance: 0.72) and four for aspect (Figure~\ref{fig:lda_aspect_dirs}, explained variance: 0.70). They span a convex region in both cases, suggesting that the feature vectors define structured subspaces. See App.~\ref{app:cluster_quality} for additional cluster quality metrics of both models.

\begin{figure}[!h]
    \centering
    \includegraphics[width=1\linewidth]{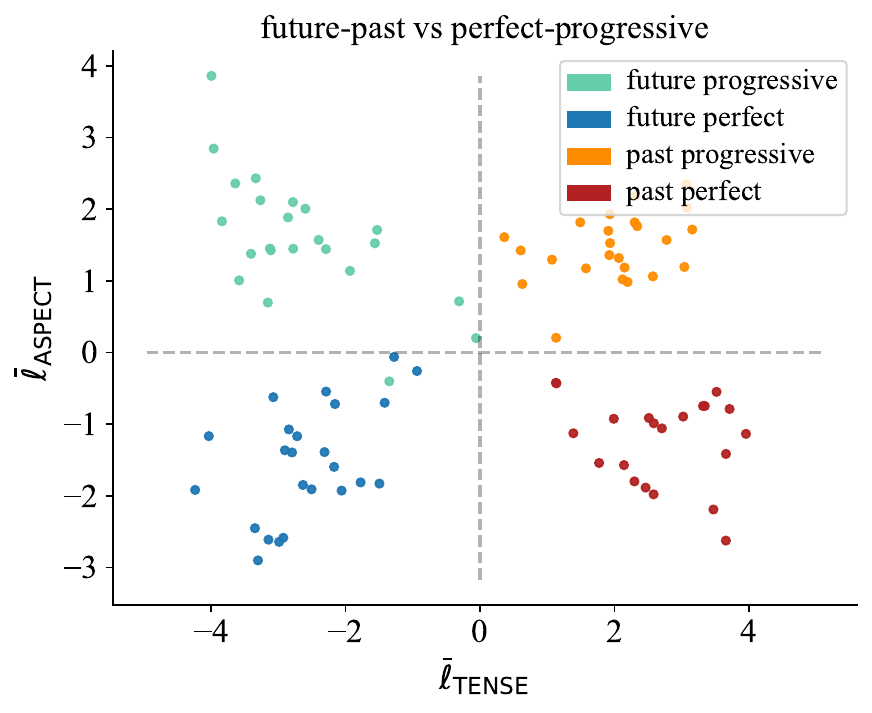}
    \caption{Projection of Llama-8B hidden states (L7) along $\bar{\ell}_{\text{\scriptsize\textsc{tense}}}$ and $\bar{\ell}_{\text{\scriptsize\textsc{aspect}}}$ clearly separate the data points according to their grammatical property.}
    \label{fig:lda_contrast}
\end{figure}

\paragraph{Tense and aspect exhibit representational independence.} 
Next, we examine the relationship between the two grammatical categories by projecting representations onto their respective latent dimensions. Specifically, we use the binary contrasts $\bar{\ell}_{\text{\scriptsize\textsc{tense}}} = \bar{\ell}_{\text{\scriptsize\textsc{future}}}-\bar{\ell}_{\text{\scriptsize\textsc{past}}}$ and $\bar{\ell}_{\text{\scriptsize\textsc{aspect}}} = \bar{\ell}_{\text{\scriptsize\textsc{progressive}}}-\bar{\ell}_{\text{\scriptsize\textsc{perfect}}}$ as proxies for the tense and aspect dimensions\footnote{Any binary contrast among subcategories (e.g., \textsc{future-past}, \textsc{present-past}, \textsc{future-present}) lies in the same one-dimensional parent-contrast subspace. The vector differences in Fig.~\ref{fig:lda_contrast} were chosen as examples.}.

Figure~\ref{fig:lda_contrast} shows the projection of data points onto the inferred tense (x-axis) and aspect (y-axis) dimensions. Similar to earlier feature-wise visualizations, the points cluster according to their grammatical categories. The groups are organized near-orthogonally in latent space, reflecting a clear separation between tense and aspect. This is further supported by the near-zero cosine similarity of 0.02 between the vectors $\bar{\ell}_{\text{\scriptsize\textsc{tense}}}$ and $\bar{\ell}_{\text{\scriptsize\textsc{aspect}}}$ (see App.~\ref{app:cluster_quality}).

\section{Multi-Token Steering}
\label{sec:multitoken_steering}
After identifying correlational evidence for tense and aspect directions in model representations, we test their causal impact on model behavior via targeted interventions during text generation. We evaluate functional selectivity by checking whether the manipulated outputs express the steered verb property while preserving other verb features and the original meaning. This quantitative analysis is followed by a qualitative study in which we manually examine model outputs and investigate the impact of steering location, strength and duration.

\subsection{Experimental Setup} 
\paragraph{Tasks.} We consider three complementary tasks (prompt details in App.~\ref{app:testset_steering}):

\begin{enumerate}[leftmargin=*]
    \item \textbf{Random Sentence Task:} The model is prompted to generate an open-ended sentence, testing whether grammatical concepts can be induced in semantically unconstrained settings.
    \item \textbf{Repetition Task:} In a few-shot setup, the model must repeat a sentence after two example repetitions. This copying task requires interventions to override contextual information. It evaluates the ability to steer generation when the model's default behavior follows pattern induction.
    \item \textbf{Temporal Translation Task:} We use a similar few-shot setup where the model must "translate" a sentence into a different tense or aspect. Unlike repetition, this requires internal transformation, allowing us to test whether interventions can influence more complex linguistic transformations.

\end{enumerate}

\paragraph{Steering Methods.} We perform steering at a single transformer layer $l$ on the final position $i=-1$ of the input sequence at every generation step. Concretely, we update the residual stream activation vector $h_i^l \in \mathbb{R}^d$ by adding and/or subtracting the normalized LDA-derived concept directions from earlier. These unit vectors correspond to specific tense and aspect values. A scalar steering factor $\alpha \in \mathbb{R}$ scales the strength of the modification. We evaluate three distinct steering strategies across all layers and different $\alpha$ values:

\begin{enumerate}[leftmargin=*]
    \item \textbf{Target-Addition Only (TA):} The standard steering approach which is commonly used in related work such as \citet{rimsky-etal-2024-steering},  directly adds the normalized target direction $\bar{\ell}_T$ to the current activation: 
    \begin{equation} h^{\textrm{steered}} = h_i^l + \alpha \bar{\ell}_T. \end{equation}
    \item \textbf{Target-Addition with Source-Subtraction (TA+SS):} To simultaneously steer the target concept and suppress a known source concept $\bar{\ell}_S$, we introduce source-subtraction, which subtracts the source direction with equal weight. This is particularly useful for non-binary features, where source and target are not simply inverses:
    \begin{equation} h^{\textrm{steered}} = h_i^l + \alpha \bar{\ell}_T - \alpha \bar{\ell}_S.
    \end{equation}
    \item \textbf{Target-Addition with Projection Subtraction (TA+Proj-SS):} Instead of subtracting the full source vector, this method removes only the component of the activation that lies along $\bar{\ell}_S$. This is achieved by computing and subtracting the projection of $h_i^l$ onto $\bar{\ell}_S$:
    \begin{equation} h^{\textrm{steered}} = h_i^l + \alpha \bar{\ell}_T - (h_i^l \cdot \bar{\ell}_S)\bar{\ell}_S.
    \end{equation}
\end{enumerate}

To ensure comparability between steered and original generations, we use greedy decoding, where the most likely token is selected at each step. 

\paragraph{Evaluation Metrics.} We evaluate steering success by giving the generated outputs to the model in a new forward pass without any interventions, extracting their representations and applying the trained probing classifiers, following \citet{brinkmann2025large}. We probe not only for the steering target but also for the respective other property. To quantify the effect of our interventions, we define the following four performance metrics:
\begin{align*}
\text{Steering Success} 
&= \frac{\lvert S\rvert}{N}, \\
\text{Degenerate Rate} 
&= \frac{\lvert D\rvert}{N},\\
  \text{Efficacy} 
    &= \frac{\lvert S \setminus D\rvert}{N}, \\
 \text{Selectivity}
 &= \frac{\lvert S_{F} \setminus D \rvert}{N}.
\end{align*}

\noindent Here, $N$ is the number of test samples\footnote{$N$ is task dependent, see App.~\ref{app:testset_steering}.}, \(S\subseteq\{1,\dots,N\}\) is the set of successfully steered samples and \(D\subseteq\{1,\dots,N\}\) the set of degenerate outputs. An output is considered degenerate, if it either (i) forms an incomplete sentence by missing a verb, as detected by a part-of-speech (POS) tagger, or (ii) exhibits excessive n-gram repetition or low n-gram diversity (see App.~\ref{app:steering_metrics} for thresholds). \(S_{F}\) is the subset of \(S\) for which the probe's label for the not steered property stays constant. Finally, we report the relative change in perplexity to measure the impact of steering on fluency and coherence.

\begin{figure}[!t]
    \centering
    \includegraphics[width=1\linewidth]{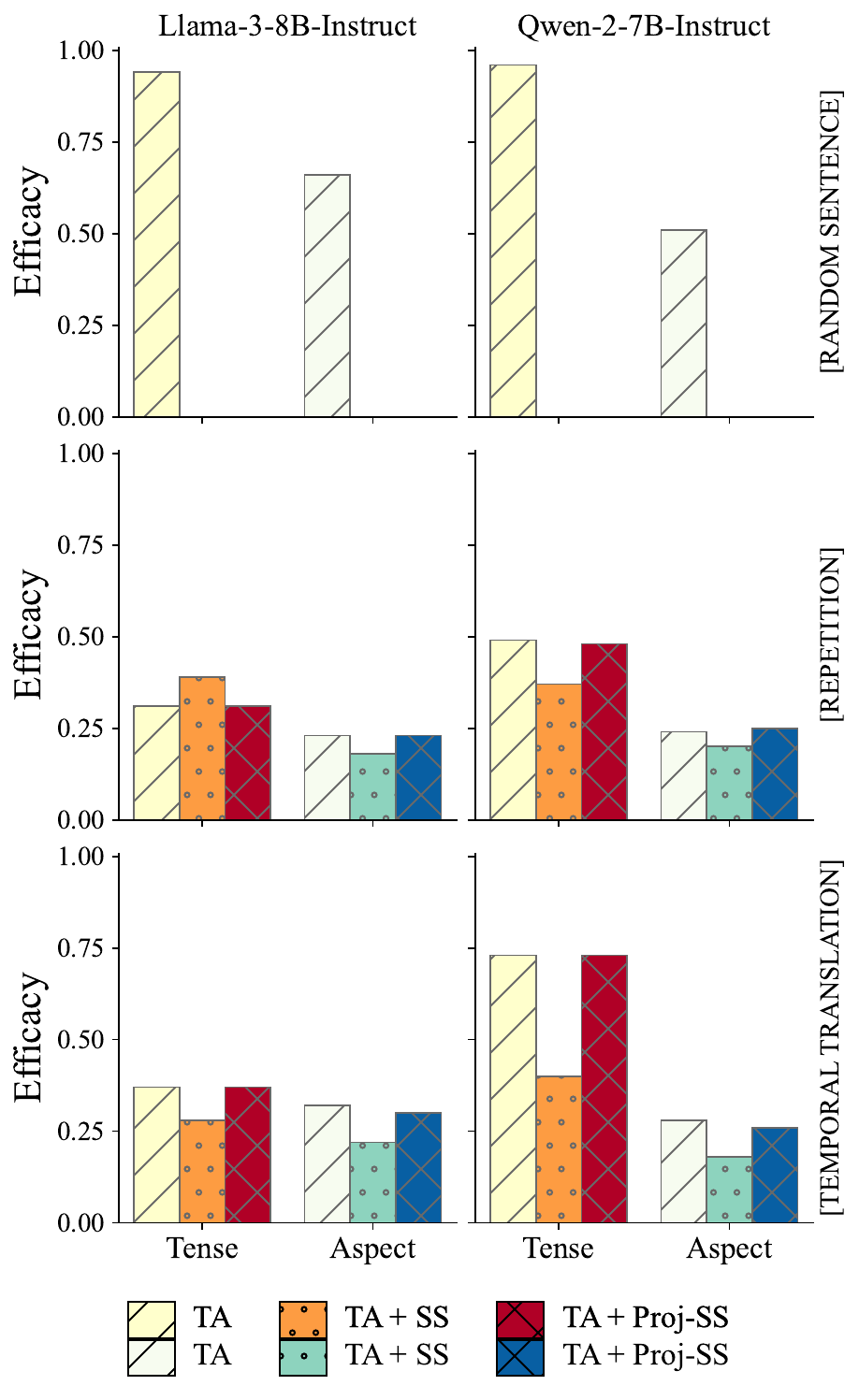}
    \caption{Efficacy of different steering methods. TA+SS and TA+Proj-SS are not applied for the Random Sentence Task because there is no source feature direction to be subtracted. Both models show similar trends with tense being easier to steer than aspect, and random sentences easier than few-shot tasks.}
    \label{fig:steering_success}
\end{figure}

\subsection{Quantitative Results}
\label{sec:quantitative_results}
For each task and steering method, we perform a grid search over all layers and selected $\alpha$ values (see App.~\ref{app:gridsearch_steering}) and report the configuration that yields the highest efficacy (Figure~\ref{fig:steering_success}).

\paragraph{Steering success varies widely by task and target.} Overall, steering tense achieves substantially higher success than aspect. For example, on the random sentence task, efficacy is near-perfect for tense (94\% for Llama-8B, 96\% for Qwen-7B) but noticeably lower for aspect (66\% for Llama-8B, 51\% for Qwen-7B). Steering in the few-shot settings reduces performance for both targets, with the best scenario reaching 73\% (Qwen-7B, tense) and the worst just 18\% (aspect for both models). Steering tense or aspect is significantly more difficult when the task requires conflicting verb properties to the steering target\footnote{E.g., for the repetition task, the sentence "I jumped into the lake." needs to be repeated in its original tense (past), but when steering "future", the intervention is contradicting to the tense the task would require.}. This trend is largely consistent across both model architectures.

Although efficacy implicitly captures output quality through the degenerate rate, we additionally report relative perplexity changes between steered and unsteered generations to evaluate the general impact of steering on text coherence and fluency. Results show low perplexity increases across most scenarios, with few outliers, demonstrating that our steering vectors maintain output quality and perform targeted interventions (App.~\ref{app:perplexity}).

\begin{figure}[!t]
    \centering
    \includegraphics[width=1\linewidth]{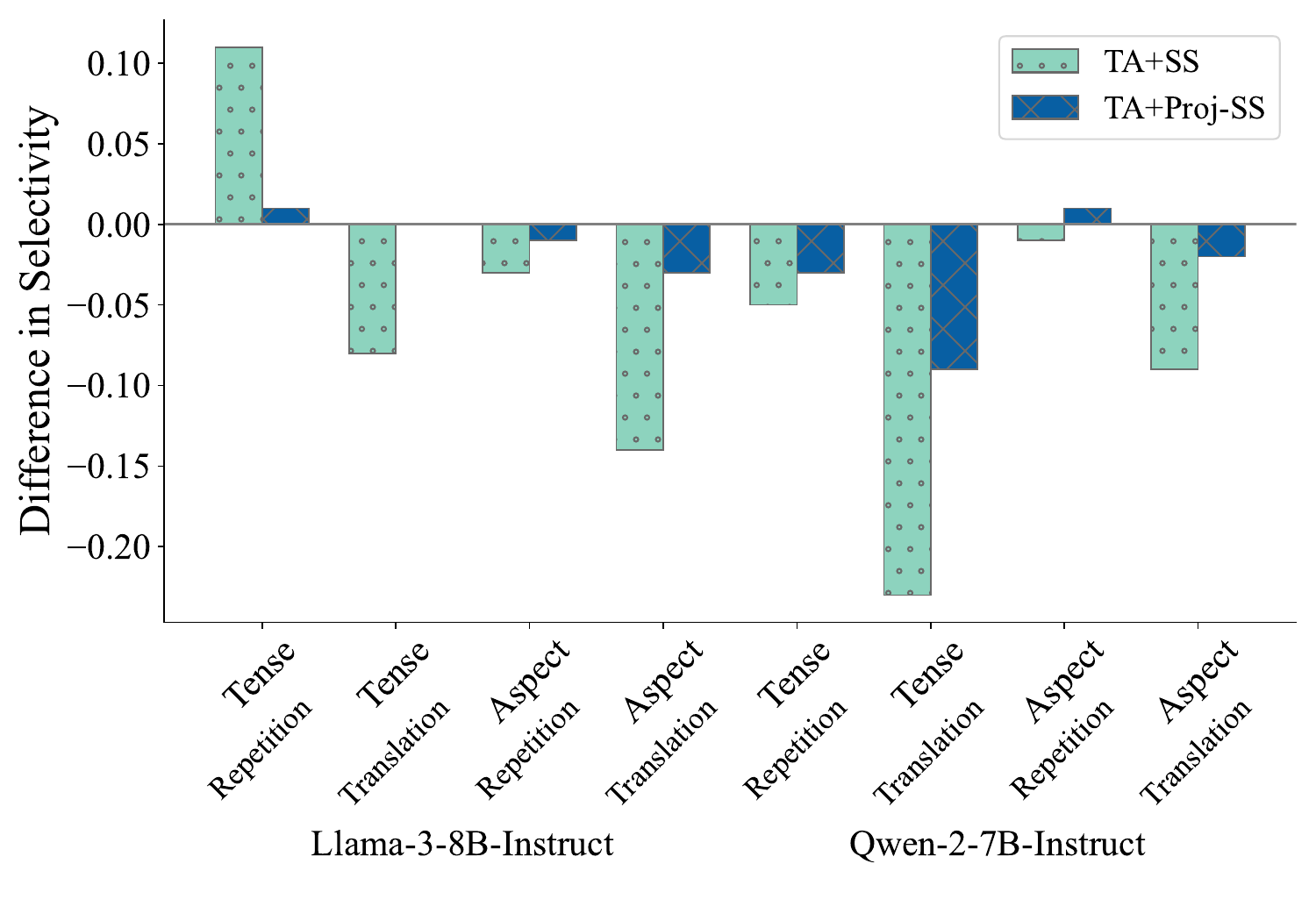}
    \caption{Subtracting the source concept vector (TA+SS) reduces selectivity. Negative bars denote selectivity drop with respect to simple target addition, but this effect can be partially mitigated through more targeted subtraction (TA+Proj-SS).}
    \label{fig:selectivity_per_steering}
    \vspace{-4mm}
\end{figure}

\paragraph{Projection subtraction improves selectivity.} Another interesting finding is that subtracting the source concept vector often harms steering performance (-4\% and -12.5\% on average for Llama-8B and Qwen-7B respectively), potentially because it introduces too much additional change to the residual stream. It also reduces selectivity (Figure~\ref{fig:selectivity_per_steering}). However, replacing full vector subtraction with projection subtraction mostly “mitigates” this issue, both for efficacy and selectivity, indicating that more targeted interventions—removing only the component aligned with the source direction—are more effective. We explore this finding further in \textsection~\ref{sec:case_study} and provide details on the correlation between efficacy and selectivity in App.~\ref{app:selectivity}.

\paragraph{Activation norm determines steering factor.} We find that Llama-8B requires significantly lower $\alpha$ values (5-25) for effective steering compared to Qwen-7B (100-250), a pattern consistent across tasks. Moreover, the optimal $\alpha$ tends to increase with depth for both models (App.~\ref{app:activation_norm}). One explanation lies in the activation norm, which increases similarly across layers and is generally higher for Qwen-7B. \citet{kobayashi-etal-2020-attention} have shown that layers with larger activation norms carry more information, suggesting that stronger interventions are required to overwrite pre-existing signals. To test this hypothesis, we examine the projection magnitude onto the source feature direction, that is, the strength with which the original tense and aspect are encoded, and observe a similar growth with depth. Thus, even though the feature's strength stays roughly constant relative to the activation norm, its absolute magnitude grows across layers, requiring proprotionally larger $\alpha$ values for effective steering.

\subsection{Qualitative Behavior and Failure Modes}
We present example outputs for the random sentence task in Table~\ref{tab:steer_random_examples} and for the few-shot tasks in Table~\ref{tab:repetition_examples} and App.~\ref{app:steering_examples_tt}. They demonstrate that while steering grammatical properties is possible, it can also lead to unintended changes in content---from slight alterations (e.g., in the repetition task) to complete topic shifts (e.g., in the random sentence task). This undesired behavior is not captured by the four evaluation metrics, but is crucial to identify when aiming for targeted and selective steering.

To assess topic shift across tasks, we focus on the most effective steering method for each model and target. We compute the semantic similarity metric BERTScore \citep{zhang2020bertscore} between unsteered and steered outputs, considering only successfully steered samples. On average, steering during the few-shot tasks causes less topic shift (i.e., higher semantic similarity in Table~\ref{tab:topic_shift}), likely because guided generation constrains their output and maintains consistency. In contrast, open-ended generation leaves more room for the steering intervention to reshape the output, leading to higher inconsistency.

\begin{table}[h]
    \centering
    \small
    \setlength{\tabcolsep}{3pt} 
    \begin{tabular}{ccc}
         \textbf{Random Sentence} & \textbf{Repetition} & \textbf{Temporal Translation}\\
         \midrule
         $0.56 \pm 0.06$ & $0.69 \pm 0.09$ & $0.77 \pm 0.07$ 
    \end{tabular}
    \caption{BERTScore between unsteered and steered output, averaged across models and targets, reveals higher consistency for few-shot tasks.}
    \label{tab:topic_shift}
\end{table}

\begin{table*}[!t]
    \centering
    \begin{tabular}{p{1.65cm}|p{4.2cm} p{4.2cm} p{4.2cm}}
    \toprule
    & \multicolumn{3}{c}{\textsc{Prompt}} \\
    \cmidrule{2-4}
        \textsc{Target} & \textbf{\small{Generate a single sentence:}} & \textbf{\small{Output a random sentence:}} & \textbf{\small{Generate one sentence:}} \\
        \midrule
        \textsc{-} & 
        \colorbox{cyan!25}{\begin{minipage}{4.0cm} \small The sun was setting over the ocean, casting a warm golden light over the waves.\end{minipage}} & 
        \colorbox{orange!25} {\begin{minipage}{4cm}
        \small The sun is shining brightly in the sky.\end{minipage}} & 
        \colorbox{purple!25} {\begin{minipage}{4cm}
        \small The new policy will require all employees to undergo a background check before being hired.\end{minipage}} \\
         \midrule
        \colorbox{cyan!25} {\textsc{past}} & 
        \colorbox{cyan!25} {\begin{minipage}{4cm} \small The sun was setting over the vast, open plain, casting a warm glow over the scattered wildflowers that swayed gently in the breeze.\end{minipage}} & 
        \colorbox{cyan!25} {\begin{minipage}{4cm} \small The sun was shining brightly on the day of the wedding.\end{minipage}} & 
        \colorbox{cyan!25}  {\begin{minipage}{4cm} \small The sun was shining brightly on the day of the wedding, casting a warm glow over the happy couple as they exchanged their vows.\end{minipage}} \\
         \midrule
        \colorbox{orange!25} {\textsc{present}} & \colorbox{orange!25} {\begin{minipage}{4cm} \small I am a 25-year-old woman who has been a vegetarian for 10 years and has been a vegan for 5 years.\end{minipage}} & \colorbox{orange!25} {\begin{minipage}{4cm} \small I am a 25-year-old man who is a student at a university in the United States.\end{minipage}} &   
        \colorbox{orange!25} {\begin{minipage}{4cm} \small I am a 25-year-old woman who is a student at a university in the United States.\end{minipage}} \\
         \midrule
        \colorbox{purple!25} {\textsc{future}} & 
        \colorbox{purple!25} {\begin{minipage}{4cm} \small The new policy will be implemented in all schools by the start of the next school year.\end{minipage}} & 
        \colorbox{purple!25} {\begin{minipage}{4cm} \small I'm going to the store to buy some milk.\end{minipage}} & 
        \colorbox{purple!25} {\begin{minipage}{4cm} \small The new policy will be implemented in all schools by the end of the 2023-2024 school year.\end{minipage}} \\
        \bottomrule
    \end{tabular}
    \caption{Examples of steering Llama-8B. Steering enforces the target tense regardless of the model's original choice, but often shifts the topic. Colors indicate tense, identified via probing classifiers.}
    \label{tab:steer_random_examples}
\end{table*}

\subsection{Exploring the Effect of Steering Location}
\label{sec:case_study}
Building on the finding that optimal steering strength is model- and layer-dependent, we now explore where and for how long to apply steering to maximize efficacy and reduce side effects. We conduct a qualitative case study, targeting two verb tenses across three examples per few-shot task using TA and TA+SS. Due to the need for token-level POS-annotations, we limit our analysis to a small set of examples. 

We differentiate two steering locations: the \textit{prompt} (i.e., modifying input representations during the initial forward pass), and the \textit{generated tokens} (i.e., intervening during subsequent generation steps). For each, we compare single versus multi-token interventions and both index-based and POS-informed steering. We find that the same steering vector can lead to successful modification, no effect or even degeneration on the same sample, depending on the steering location. Representative outputs are shown in Tables~\ref{tab:repetition_pos} and App.~\ref{app:steering_examples_tt}.

\begin{table}
\small
\begin{tabular}{l|l}
 \toprule
 \textbf{Prompt} & \begin{tabular}[t]{@{}l@{}}
Maya was writing a story. \textbackslash\textbackslash{} \\
Maya was writing a story. \\\\
She accepted that offer.\textbackslash\textbackslash{} \\
She accepted that offer. \\\\
\colorbox{orange!25}{He has thought about this.} \textbackslash\textbackslash{} 
\end{tabular} \\
\midrule
\textbf{Output (unsteered)} & 
\colorbox{orange!25}{He has thought about this.}  \\
\midrule
\textbf{Output (steered \colorbox{cyan!25}{\textit{past}}}) & \begin{tabular}[t]{@{}l@{}}
\colorbox{cyan!25}{He had not thought about} \\ \colorbox{cyan!25}{anything else.} \end{tabular} \\
\bottomrule
\end{tabular}
\caption{Example of Llama-8B on repetition task. \textit{Orange}: tense of unsteered outputs, \textit{blue}: target.}
\label{tab:repetition_examples}
\end{table}

\begin{table*}[!t]
\small
\setlength{\tabcolsep}{4pt} 
\begin{tabular}{lllll}
\toprule
 \textbf{Steering Position} & \textbf{Prompt Tokens} & \textbf{Generated Tokens} & \textbf{Output (TA)} & \textbf{Output (TA-SS)}\\
 \midrule
 all verb tokens in prompt & ...\texttt{It} \colorbox{orange!25}{\texttt{is snow ing}} \texttt{.} \texttt{\textbackslash\textbackslash} &  \texttt{It} \texttt{is} \texttt{snow} \texttt{ing} \texttt{.} & \textcolor{green!60!black}{It was snowing.} & \textcolor{green!60!black}{It was snowing.}\\
 last verb token in prompt & ...\texttt{It} \texttt{is} \texttt{snow} \colorbox{orange!25}{\texttt{ing}} \texttt{.} \texttt{\textbackslash\textbackslash} &  \texttt{It} \texttt{is} \texttt{snow} \texttt{ing} \texttt{.} & It is snowing. & It is snowing. \\
   sentence end in prompt & ...\texttt{It} \texttt{is} \texttt{snow} \texttt{ing}\colorbox{orange!25}{\texttt{.}} \texttt{\textbackslash\textbackslash} &  \texttt{It} \texttt{is} \texttt{snow} \texttt{ing} \texttt{.}& It is snowing. & It is snowing.\\
   final token in prompt & ...\texttt{It} \texttt{is} \texttt{snow} \texttt{ing} \texttt{.} \colorbox{orange!25}{\texttt{\textbackslash\textbackslash}} &  \texttt{It} \texttt{is} \texttt{snow} \texttt{ing} \texttt{.} & It is snowing. & It is snowing.\\
final tokens during generation & ...\texttt{It} \texttt{is} \texttt{snow} \texttt{ing} \texttt{.} \colorbox{orange!25}{\texttt{\textbackslash\textbackslash}} &  \colorbox{orange!25}{\texttt{It}} \colorbox{orange!25}{\texttt{is}} \colorbox{orange!25}{\texttt{snow}} \colorbox{orange!25}{\texttt{ing}} \texttt{.} & \textcolor{purple!80}{It was a day...} & \textcolor{purple!80}{It was a long...}\\
  generated token before verb & ...\texttt{It} \texttt{is} \texttt{snow} \texttt{ing} \texttt{.} \texttt{\textbackslash\textbackslash} &  \colorbox{orange!25}{\texttt{It}} \texttt{is} \texttt{snow} \texttt{ing} \texttt{.} & \textcolor{green!60!black}{It was snowing.} & \textcolor{green!60!black}{It was snowing.} \\
 first generated verb token & ...\texttt{It} \texttt{is} \texttt{snow} \texttt{ing} \texttt{.} \texttt{\textbackslash\textbackslash} &  \texttt{It} \colorbox{orange!25}{\texttt{is}} \texttt{snow} \texttt{ing} \texttt{.} & It is snowing. & It is snowing. \\
 all generated verb token & ...\texttt{It} \texttt{is} \texttt{snow} \texttt{ing} \texttt{.} \texttt{\textbackslash\textbackslash} &  \texttt{It} \colorbox{orange!25}{\texttt{is snow ing}} \texttt{.} & It is snowing. & It is snowing. \\
  \bottomrule
 \end{tabular}
 \caption{Llama-8B: Steering \textsc{past} on the repetition task for the prompt: "He is crying. \textbackslash\textbackslash{} He is crying.\texttt{\textbackslash n}\texttt{\textbackslash n} We were having dinner. \textbackslash\textbackslash{} We were having dinner. \texttt{\textbackslash n}\texttt{\textbackslash n} It is snowing. \textbackslash\textbackslash{}". \textit{Generated Tokens}: unsteered output.}
  \label{tab:repetition_pos}
 \end{table*}

\paragraph{Generation-time steering is more effective.} Steering during generation is consistently more effective across both tasks, while prompt-based interventions succeed only for the repetition task. A plausible explanation is that repetition relies on pattern induction, whereas temporal translation involves more complex grammatical reasoning, making earlier interventions harder.
\paragraph{Steering before the verb works best.} The optimal steering duration depends on the location. For effective prompt interventions, all verb tokens need to be steered---suggesting that, while multi-token expressions may aggregate meaning at their final token \citep{feucht-etal-2024-token}, a single impulse at the end of a verb phrase is insufficient to overwrite all previous verb information. In contrast, generation-time steering is sensitive to timing and duration: late or extended steering can cause topic shift and degeneration. We find that verb properties are steered most effectively just before the generated verb.

\paragraph{Target addition does not require source subtraction.} Interestingly, steering positions that succeed with TA+SS also succeed with TA, which aligns with the findings of our quantitative method comparison in \textsection~\ref{sec:quantitative_results}. If the source and target directions were truely orthogonal, then removing the source should improve steering by eliminating conflicting information about the feature of interest. The comparable results between TA and TA+SS, however, suggest that the directions may not be fully independent, or that the effect of the target direction dominates in practice. Understanding how these directions interact---and whether source subtraction helps or hinders---remains an open question for future work.

While our analysis is qualitative and grounded in grammatical insights into sentence structure of prompt and output, it emphasizes the critical role of steering position and duration, in addition to steering strength. We encourage future research to develop automated methods for identifying task-specific optimal steering positions, building on early work such as \citet{lee2025programming}.

\section{Discussion and Conclusion}
\label{sec:discussion_conclusion}
\paragraph{Syntactic categories are represented orthogonally in latent space.} Our findings show that language models obtain structural representations of tense and aspect that go beyond surface-level pattern recognition. They can be probed and visualized, showing structural organization of individual categories similar to those humans use to differentiate these verb properties. Values within the same grammatical category (e.g., past, present for tense) form approximately orthogonal directions in latent space. Similarly, broader tense and aspect vectors appear orthogonal to each other, highlighting their representational independence. We find that this encoding of syntactic structure has causal relevance and can be used to steer multi-token generation across different tasks.

\paragraph{Steering works for causal verification but is not a perfect method for model control yet.} We use these verb properties to study factors influencing successful steering. While our results show that steering grammatical features can work, there are pitfalls such as topic shift and output degeneration that need to be monitored. Simple metrics such as n-gram statistics, perplexity, POS-tagging and BERTScore help to track side effects, while more expensive methods (e.g., LLM-as-a-judge) can be applied for final evaluations after tuning hyperparameters.
Our results suggest that activation norm can be a useful heuristic when adjusting scaling factors, with higher norms requiring stronger steering. Further, the question of where and how long to intervene is task-dependent but can significantly affect success. We find interventions during generation to be generally more effective than steering via updating the prompt representation, particularly in cases where the target property conflicts with the task context.
For categorical target values, we find that adding the desired property vector is sufficient, removing the currently present category vector yields no additional benefit due to the approximate orthogonality of the categorical vectors.
We encourage future work to more systematically monitor side effects of steering, and to explore automated methods for optimizing steering conditions.

\section*{Acknowledgments}
The authors acknowledge support by the state of Baden-Württemberg through bwHPC
and the German Research Foundation (DFG) through grant INST 35/1597-1 FUGG.

\section*{Limitations}
\paragraph{Temporal Expression in Language.} Our study focuses on the two verbal properties tense and aspect in English, which has a very regular inflection system, to investigate categorical and combinatorial grammatical structures in language models. While we find evidence that model representations reflect human-like grammar organization, our study has some limitations. First, we restrict our analysis to sentences containing a single unique tense-aspect combination. This is a simplification, as natural language frequently expresses complex temporal relations and event succession through multiple verb phrases with different tense-aspect combinations. Second, as noted by \citet{klein2009time}, tense and aspect are only two of six known strategies for expressing temporal information in language. Other mechanisms such as temporal adverbs are used especially in tenseless languages. We leave it to future work to investigate mechanism-independent time representations (e.g., consistent representation of "past" in different surface forms, such as "he was" and "yesterday"), as well as language-independent tense representations (e.g., cross-lingual past-present contrast, see \citet{brinkmann2025large}).

\paragraph{Scope of Steering Analysis.}  
Our analysis focuses on steering grammatical properties in three tasks, which provide a controlled testbed for comparing different types of interventions. However, side effects such as topic shift may manifest differently when steering other concepts like sentiment. Expanding the range of tasks and steering targets could help identify more general conditions for effective steering, which provides a promising direction for future work.

\bibliography{anthology,custom}

\begin{thebibliography}{57}
\providecommand{\natexlab}[1]{#1}

\bibitem[{Arora et~al.(2024)Arora, Jurafsky, and Potts}]{arora2024causalgym}
Aryaman Arora, Dan Jurafsky, and Christopher Potts. 2024.
\newblock \href {https://doi.org/10.18653/v1/2024.acl-long.785} {{C}ausal{G}ym: Benchmarking causal interpretability methods on linguistic tasks}.
\newblock In \emph{Proceedings of the 62nd Annual Meeting of the Association for Computational Linguistics (Volume 1: Long Papers)}, pages 14638--14663, Bangkok, Thailand. Association for Computational Linguistics.

\bibitem[{Belinkov(2022)}]{belinkov-2022-probing}
Yonatan Belinkov. 2022.
\newblock \href {https://doi.org/10.1162/coli_a_00422} {Probing classifiers: Promises, shortcomings, and advances}.
\newblock \emph{Computational Linguistics}, 48(1):207--219.

\bibitem[{Bricken et~al.(2023)Bricken, Templeton, Batson, Chen, Jermyn, Conerly, Turner, Anil, Denison, Askell, Lasenby, Wu, Kravec, Schiefer, Maxwell, Joseph, Hatfield-Dodds, Tamkin, Nguyen, McLean, Burke, Hume, Carter, Henighan, and Olah}]{bricken2023monosemanticity}
Trenton Bricken, Adly Templeton, Joshua Batson, Brian Chen, Adam Jermyn, Tom Conerly, Nick Turner, Cem Anil, Carson Denison, Amanda Askell, Robert Lasenby, Yifan Wu, Shauna Kravec, Nicholas Schiefer, Tim Maxwell, Nicholas Joseph, Zac Hatfield-Dodds, Alex Tamkin, Karina Nguyen, and 6 others. 2023.
\newblock Towards monosemanticity: Decomposing language models with dictionary learning.
\newblock \emph{Transformer Circuits Thread}.
\newblock Https://transformer-circuits.pub/2023/monosemantic-features/index.html.

\bibitem[{Brinkmann et~al.(2025)Brinkmann, Wendler, Bartelt, and Mueller}]{brinkmann2025large}
Jannik Brinkmann, Chris Wendler, Christian Bartelt, and Aaron Mueller. 2025.
\newblock \href {https://aclanthology.org/2025.naacl-long.312/} {Large language models share representations of latent grammatical concepts across typologically diverse languages}.
\newblock In \emph{Proceedings of the 2025 Conference of the Nations of the Americas Chapter of the Association for Computational Linguistics: Human Language Technologies (Volume 1: Long Papers)}, pages 6131--6150, Albuquerque, New Mexico. Association for Computational Linguistics.

\bibitem[{Chen et~al.(2024)Chen, Shwartz-Ziv, Cho, Leavitt, and Saphra}]{chen2024sudden}
Angelica Chen, Ravid Shwartz-Ziv, Kyunghyun Cho, Matthew~L Leavitt, and Naomi Saphra. 2024.
\newblock \href {https://openreview.net/forum?id=MO5PiKHELW} {Sudden drops in the loss: Syntax acquisition, phase transitions, and simplicity bias in {MLM}s}.
\newblock In \emph{The Twelfth International Conference on Learning Representations}.

\bibitem[{Conneau et~al.(2018)Conneau, Kruszewski, Lample, Barrault, and Baroni}]{conneau2018cram}
Alexis Conneau, German Kruszewski, Guillaume Lample, Lo{\"i}c Barrault, and Marco Baroni. 2018.
\newblock \href {https://doi.org/10.18653/v1/P18-1198} {What you can cram into a single {\$}{\&}!{\#}* vector: Probing sentence embeddings for linguistic properties}.
\newblock In \emph{Proceedings of the 56th Annual Meeting of the Association for Computational Linguistics (Volume 1: Long Papers)}, pages 2126--2136, Melbourne, Australia. Association for Computational Linguistics.

\bibitem[{Diego~Simon et~al.(2024)Diego~Simon, d'Ascoli, Chemla, Lakretz, and King}]{diego2024polar}
Pablo~J Diego~Simon, St{\'e}phane d'Ascoli, Emmanuel Chemla, Yair Lakretz, and Jean-R{\'e}mi King. 2024.
\newblock A polar coordinate system represents syntax in large language models.
\newblock \emph{Advances in Neural Information Processing Systems}, 37:105375--105396.

\bibitem[{Dubey et~al.(2024)Dubey, Jauhri, Pandey, Kadian, Al-Dahle, Letman, Mathur, Schelten, Yang, Fan, Goyal, Hartshorn, Yang, Mitra, Sravankumar, Korenev, Hinsvark, Rao, Zhang, Rodriguez, Gregerson, Spataru, Rozière, Biron, Tang, Chern, Caucheteux, Nayak, Bi, Marra, McConnell, Keller, Touret, Wu, Wong, Ferrer, Nikolaidis, Allonsius, Song, Pintz, Livshits, Esiobu, Choudhary, Mahajan, Garcia-Olano, Perino, Hupkes, Lakomkin, AlBadawy, Lobanova, Dinan, Smith, Radenovic, Zhang, Synnaeve, Lee, Anderson, Nail, Mialon, Pang, Cucurell, Nguyen, Korevaar, Xu, Touvron, Zarov, Ibarra, Kloumann, Misra, Evtimov, Copet, Lee, Geffert, Vranes, Park, Mahadeokar, Shah, van~der Linde, Billock, Hong, Lee, Fu, Chi, Huang, Liu, Wang, Yu, Bitton, Spisak, Park, Rocca, Johnstun, Saxe, Jia, Alwala, Upasani, Plawiak, Li, Heafield, Stone, and et~al.}]{grattafiori2024llama3herdmodels}
Abhimanyu Dubey, Abhinav Jauhri, Abhinav Pandey, Abhishek Kadian, Ahmad Al-Dahle, Aiesha Letman, Akhil Mathur, Alan Schelten, Amy Yang, Angela Fan, Anirudh Goyal, Anthony Hartshorn, Aobo Yang, Archi Mitra, Archie Sravankumar, Artem Korenev, Arthur Hinsvark, Arun Rao, Aston Zhang, and 82 others. 2024.
\newblock \href {https://doi.org/10.48550/arXiv.2407.21783} {The llama 3 herd of models}.
\newblock \emph{CoRR}, abs/2407.21783.

\bibitem[{Elhadad(2010)}]{elhadad-2010-book}
Michael Elhadad. 2010.
\newblock \href {https://doi.org/10.1162/coli_r_00022} {Book review: Natural language processing with python by steven bird, ewan {K}lein, and edward loper}.
\newblock \emph{Computational Linguistics}, 36(4).

\bibitem[{Elhage et~al.(2022)Elhage, Hume, Olsson, Schiefer, Henighan, Kravec, Hatfield-Dodds, Lasenby, Drain, Chen, Grosse, McCandlish, Kaplan, Amodei, Wattenberg, and Olah}]{elhage2022toy}
Nelson Elhage, Tristan Hume, Catherine Olsson, Nicholas Schiefer, Tom Henighan, Shauna Kravec, Zac Hatfield-Dodds, Robert Lasenby, Dawn Drain, Carol Chen, Roger Grosse, Sam McCandlish, Jared Kaplan, Dario Amodei, Martin Wattenberg, and Christopher Olah. 2022.
\newblock Toy models of superposition.
\newblock \emph{Transformer Circuits Thread}.
\newblock Https://transformer-circuits.pub/2022/toy\_model/index.html.

\bibitem[{Ferrando and Costa-juss{\`a}(2024)}]{ferrando-costa-jussa-2024-similarity}
Javier Ferrando and Marta~R. Costa-juss{\`a}. 2024.
\newblock \href {https://doi.org/10.18653/v1/2024.findings-emnlp.591} {On the similarity of circuits across languages: a case study on the subject-verb agreement task}.
\newblock In \emph{Findings of the Association for Computational Linguistics: EMNLP 2024}, pages 10115--10125, Miami, Florida, USA. Association for Computational Linguistics.

\bibitem[{Feucht et~al.(2024)Feucht, Atkinson, Wallace, and Bau}]{feucht-etal-2024-token}
Sheridan Feucht, David Atkinson, Byron~C Wallace, and David Bau. 2024.
\newblock \href {https://doi.org/10.18653/v1/2024.emnlp-main.543} {Token erasure as a footprint of implicit vocabulary items in {LLM}s}.
\newblock In \emph{Proceedings of the 2024 Conference on Empirical Methods in Natural Language Processing}, pages 9727--9739, Miami, Florida, USA. Association for Computational Linguistics.

\bibitem[{Geiger et~al.(2025)Geiger, Ibeling, Zur, Chaudhary, Chauhan, Huang, Arora, Wu, Goodman, Potts, and Icard}]{geigercausalabstractions}
Atticus Geiger, Duligur Ibeling, Amir Zur, Maheep Chaudhary, Sonakshi Chauhan, Jing Huang, Aryaman Arora, Zhengxuan Wu, Noah Goodman, Christopher Potts, and Thomas Icard. 2025.
\newblock \href {http://jmlr.org/papers/v26/23-0058.html} {Causal abstraction: A theoretical foundation for mechanistic interpretability}.
\newblock \emph{Journal of Machine Learning Research}, 26(83):1--64.

\bibitem[{Gurnee and Tegmark(2024)}]{gurnee2024language}
Wes Gurnee and Max Tegmark. 2024.
\newblock \href {https://openreview.net/forum?id=jE8xbmvFin} {Language models represent space and time}.
\newblock In \emph{The Twelfth International Conference on Learning Representations}.

\bibitem[{He et~al.(2024)He, Chen, Nie, Li, and Brennan}]{he-etal-2024-decoding}
Linyang He, Peili Chen, Ercong Nie, Yuanning Li, and Jonathan~R. Brennan. 2024.
\newblock \href {https://aclanthology.org/2024.lrec-main.402/} {Decoding probing: Revealing internal linguistic structures in neural language models using minimal pairs}.
\newblock In \emph{Proceedings of the 2024 Joint International Conference on Computational Linguistics, Language Resources and Evaluation (LREC-COLING 2024)}, pages 4488--4497, Torino, Italia. ELRA and ICCL.

\bibitem[{Heinzerling and Inui(2024)}]{heinzerling2024monotonic}
Benjamin Heinzerling and Kentaro Inui. 2024.
\newblock \href {https://doi.org/10.18653/v1/2024.acl-short.18} {Monotonic representation of numeric attributes in language models}.
\newblock In \emph{Proceedings of the 62nd Annual Meeting of the Association for Computational Linguistics (Volume 2: Short Papers)}, pages 175--195, Bangkok, Thailand. Association for Computational Linguistics.

\bibitem[{Hewitt and Manning(2019)}]{hewitt-manning-2019-structural}
John Hewitt and Christopher~D. Manning. 2019.
\newblock \href {https://doi.org/10.18653/v1/N19-1419} {{A} structural probe for finding syntax in word representations}.
\newblock In \emph{Proceedings of the 2019 Conference of the North {A}merican Chapter of the Association for Computational Linguistics: Human Language Technologies, Volume 1 (Long and Short Papers)}, pages 4129--4138, Minneapolis, Minnesota. Association for Computational Linguistics.

\bibitem[{Honnibal et~al.(2020)Honnibal, Montani, Van~Landeghem, Boyd et~al.}]{honnibal2020spacy}
Matthew Honnibal, Ines Montani, Sofie Van~Landeghem, Adriane Boyd, and 1 others. 2020.
\newblock spacy: Industrial-strength natural language processing in python.

\bibitem[{Hu et~al.(2020)Hu, Gauthier, Qian, Wilcox, and Levy}]{hu2020systematic}
Jennifer Hu, Jon Gauthier, Peng Qian, Ethan Wilcox, and Roger Levy. 2020.
\newblock \href {https://doi.org/10.18653/v1/2020.acl-main.158} {A systematic assessment of syntactic generalization in neural language models}.
\newblock In \emph{Proceedings of the 58th Annual Meeting of the Association for Computational Linguistics}, pages 1725--1744, Online. Association for Computational Linguistics.

\bibitem[{Huben et~al.(2024)Huben, Cunningham, Smith, Ewart, and Sharkey}]{huben2024sparse}
Robert Huben, Hoagy Cunningham, Logan~Riggs Smith, Aidan Ewart, and Lee Sharkey. 2024.
\newblock \href {https://openreview.net/forum?id=F76bwRSLeK} {Sparse autoencoders find highly interpretable features in language models}.
\newblock In \emph{The Twelfth International Conference on Learning Representations}.

\bibitem[{Hurst et~al.(2024)Hurst, Lerer, Goucher, Perelman, Ramesh, Clark, Ostrow, Welihinda, Hayes, Radford et~al.}]{hurst2024gpt}
Aaron Hurst, Adam Lerer, Adam~P Goucher, Adam Perelman, Aditya Ramesh, Aidan Clark, AJ~Ostrow, Akila Welihinda, Alan Hayes, Alec Radford, and 1 others. 2024.
\newblock Gpt-4o system card.
\newblock \emph{arXiv preprint arXiv:2410.21276}.

\bibitem[{Jumelet et~al.(2025)Jumelet, Weissweiler, and Bisazza}]{jumelet2025multiblimp}
Jaap Jumelet, Leonie Weissweiler, and Arianna Bisazza. 2025.
\newblock Multiblimp 1.0: A massively multilingual benchmark of linguistic minimal pairs.
\newblock \emph{arXiv preprint arXiv:2504.02768}.

\bibitem[{Katinskaia and Yangarber(2024)}]{katinskaia-yangarber-2024-probing}
Anisia Katinskaia and Roman Yangarber. 2024.
\newblock \href {https://doi.org/10.18653/v1/2024.findings-naacl.212} {Probing the category of verbal aspect in transformer language models}.
\newblock In \emph{Findings of the Association for Computational Linguistics: NAACL 2024}, pages 3347--3366, Mexico City, Mexico. Association for Computational Linguistics.

\bibitem[{Kingsbury and Palmer(2002)}]{kingsbury-palmer-2002-treebank}
Paul Kingsbury and Martha Palmer. 2002.
\newblock \href {https://aclanthology.org/L02-1283/} {From {T}ree{B}ank to {P}rop{B}ank}.
\newblock In \emph{Proceedings of the Third International Conference on Language Resources and Evaluation ({LREC}`02)}, Las Palmas, Canary Islands - Spain. European Language Resources Association (ELRA).

\bibitem[{Klafka and Ettinger(2020)}]{klafka2020spying}
Josef Klafka and Allyson Ettinger. 2020.
\newblock \href {https://doi.org/10.18653/v1/2020.acl-main.434} {Spying on your neighbors: Fine-grained probing of contextual embeddings for information about surrounding words}.
\newblock In \emph{Proceedings of the 58th Annual Meeting of the Association for Computational Linguistics}, pages 4801--4811, Online. Association for Computational Linguistics.

\bibitem[{Klein(2009)}]{klein2009time}
Wolfgang Klein. 2009.
\newblock \emph{How time is encoded}.
\newblock Mouton de Gruyter.

\bibitem[{Kobayashi et~al.(2020)Kobayashi, Kuribayashi, Yokoi, and Inui}]{kobayashi-etal-2020-attention}
Goro Kobayashi, Tatsuki Kuribayashi, Sho Yokoi, and Kentaro Inui. 2020.
\newblock \href {https://doi.org/10.18653/v1/2020.emnlp-main.574} {Attention is not only a weight: Analyzing transformers with vector norms}.
\newblock In \emph{Proceedings of the 2020 Conference on Empirical Methods in Natural Language Processing (EMNLP)}, pages 7057--7075, Online. Association for Computational Linguistics.

\bibitem[{Lee et~al.(2025)Lee, Padhi, Ramamurthy, Miehling, Dognin, Nagireddy, and Dhurandhar}]{lee2025programming}
Bruce~W. Lee, Inkit Padhi, Karthikeyan~Natesan Ramamurthy, Erik Miehling, Pierre Dognin, Manish Nagireddy, and Amit Dhurandhar. 2025.
\newblock \href {https://openreview.net/forum?id=Oi47wc10sm} {Programming refusal with conditional activation steering}.
\newblock In \emph{The Thirteenth International Conference on Learning Representations}.

\bibitem[{Li et~al.(2023)Li, Holtzman, Fried, Liang, Eisner, Hashimoto, Zettlemoyer, and Lewis}]{li-etal-2023-contrastive}
Xiang~Lisa Li, Ari Holtzman, Daniel Fried, Percy Liang, Jason Eisner, Tatsunori Hashimoto, Luke Zettlemoyer, and Mike Lewis. 2023.
\newblock \href {https://doi.org/10.18653/v1/2023.acl-long.687} {Contrastive decoding: Open-ended text generation as optimization}.
\newblock In \emph{Proceedings of the 61st Annual Meeting of the Association for Computational Linguistics (Volume 1: Long Papers)}, pages 12286--12312, Toronto, Canada. Association for Computational Linguistics.

\bibitem[{Logeswaran et~al.(2018)Logeswaran, Lee, and Bengio}]{logeswaran2018content}
Lajanugen Logeswaran, Honglak Lee, and Samy Bengio. 2018.
\newblock Content preserving text generation with attribute controls.
\newblock \emph{Advances in Neural Information Processing Systems}, 31.

\bibitem[{L{\'o}pez-Otal et~al.(2025)L{\'o}pez-Otal, Gracia, Bernad, Bobed, Pitarch-Ballesteros, and Angl{\'e}s-Herrero}]{lopez2025linguistic}
Miguel L{\'o}pez-Otal, Jorge Gracia, Jordi Bernad, Carlos Bobed, Luc{\'\i}a Pitarch-Ballesteros, and Emma Angl{\'e}s-Herrero. 2025.
\newblock Linguistic interpretability of transformer-based language models: a systematic review.
\newblock \emph{arXiv preprint arXiv:2504.08001}.

\bibitem[{Marcus et~al.(1993)Marcus, Santorini, and Marcinkiewicz}]{marcus-etal-1993-building}
Mitchell~P. Marcus, Beatrice Santorini, and Mary~Ann Marcinkiewicz. 1993.
\newblock \href {https://aclanthology.org/J93-2004/} {Building a large annotated corpus of {E}nglish: The {P}enn {T}reebank}.
\newblock \emph{Computational Linguistics}, 19(2):313--330.

\bibitem[{Marks and Tegmark(2024)}]{marks2024the}
Samuel Marks and Max Tegmark. 2024.
\newblock \href {https://openreview.net/forum?id=aajyHYjjsk} {The geometry of truth: Emergent linear structure in large language model representations of true/false datasets}.
\newblock In \emph{First Conference on Language Modeling}.

\bibitem[{Merullo et~al.(2024)Merullo, Eickhoff, and Pavlick}]{merullo2024language}
Jack Merullo, Carsten Eickhoff, and Ellie Pavlick. 2024.
\newblock \href {https://doi.org/10.18653/v1/2024.naacl-long.281} {Language models implement simple {W}ord2{V}ec-style vector arithmetic}.
\newblock In \emph{Proceedings of the 2024 Conference of the North American Chapter of the Association for Computational Linguistics: Human Language Technologies (Volume 1: Long Papers)}, pages 5030--5047, Mexico City, Mexico. Association for Computational Linguistics.

\bibitem[{Mueller et~al.(2024)Mueller, Brinkmann, Li, Marks, Pal, Prakash, Rager, Sankaranarayanan, Sharma, Sun, Todd, Bau, and Belinkov}]{mueller2024quest}
Aaron Mueller, Jannik Brinkmann, Millicent~L. Li, Samuel Marks, Koyena Pal, Nikhil Prakash, Can Rager, Aruna Sankaranarayanan, Arnab~Sen Sharma, Jiuding Sun, Eric Todd, David Bau, and Yonatan Belinkov. 2024.
\newblock \href {https://doi.org/10.48550/arXiv.2408.01416} {The quest for the right mediator: A history, survey, and theoretical grounding of causal interpretability}.
\newblock \emph{CoRR}, abs/2408.01416.

\bibitem[{Nanda et~al.(2023)Nanda, Lee, and Wattenberg}]{nanda-etal-2023-emergent}
Neel Nanda, Andrew Lee, and Martin Wattenberg. 2023.
\newblock \href {https://doi.org/10.18653/v1/2023.blackboxnlp-1.2} {Emergent linear representations in world models of self-supervised sequence models}.
\newblock In \emph{Proceedings of the 6th BlackboxNLP Workshop: Analyzing and Interpreting Neural Networks for NLP}, pages 16--30, Singapore. Association for Computational Linguistics.

\bibitem[{Palmer et~al.(2005)Palmer, Gildea, and Kingsbury}]{palmer-etal-2005-proposition}
Martha Palmer, Daniel Gildea, and Paul Kingsbury. 2005.
\newblock \href {https://doi.org/10.1162/0891201053630264} {The {P}roposition {B}ank: An annotated corpus of semantic roles}.
\newblock \emph{Computational Linguistics}, 31(1):71--106.

\bibitem[{Park et~al.(2024{\natexlab{a}})Park, Choe, Jiang, and Veitch}]{park2024geometry}
Kiho Park, Yo~Joong Choe, Yibo Jiang, and Victor Veitch. 2024{\natexlab{a}}.
\newblock \href {https://openreview.net/forum?id=KXuYjuBzKo} {The geometry of categorical and hierarchical concepts in large language models}.
\newblock In \emph{ICML 2024 Workshop on Mechanistic Interpretability}.

\bibitem[{Park et~al.(2024{\natexlab{b}})Park, Choe, and Veitch}]{park2023linear}
Kiho Park, Yo~Joong Choe, and Victor Veitch. 2024{\natexlab{b}}.
\newblock The linear representation hypothesis and the geometry of large language models.
\newblock In \emph{Proceedings of the 41st International Conference on Machine Learning}, ICML'24. JMLR.org.

\bibitem[{Pedregosa et~al.(2011)Pedregosa, Varoquaux, Gramfort, Michel, Thirion, Grisel, Blondel, Prettenhofer, Weiss, Dubourg et~al.}]{pedregosa2011scikit}
Fabian Pedregosa, Ga{\"e}l Varoquaux, Alexandre Gramfort, Vincent Michel, Bertrand Thirion, Olivier Grisel, Mathieu Blondel, Peter Prettenhofer, Ron Weiss, Vincent Dubourg, and 1 others. 2011.
\newblock Scikit-learn: Machine learning in python.
\newblock \emph{Journal of machine learning research}, 12(Oct):2825--2830.

\bibitem[{Pres et~al.(2024)Pres, Ruis, Lubana, and Krueger}]{pres2024reliable}
Itamar Pres, Laura Ruis, Ekdeep~Singh Lubana, and David Krueger. 2024.
\newblock \href {https://openreview.net/forum?id=7xJcX2gbm9} {Towards reliable evaluation of behavior steering interventions in {LLM}s}.
\newblock In \emph{MINT: Foundation Model Interventions}.

\bibitem[{Pustejovsky et~al.(2006)Pustejovsky, Littman, Saur{\'\i}, and Verhagen}]{pustejovsky2006timebank}
James Pustejovsky, Jessica Littman, Roser Saur{\'\i}, and Marc Verhagen. 2006.
\newblock Timebank 1.2 documentation.
\newblock \emph{Event London, no. April}, pages 6--11.

\bibitem[{Qi et~al.(2020)Qi, Zhang, Zhang, Bolton, and Manning}]{qi-etal-2020-stanza}
Peng Qi, Yuhao Zhang, Yuhui Zhang, Jason Bolton, and Christopher~D. Manning. 2020.
\newblock \href {https://doi.org/10.18653/v1/2020.acl-demos.14} {{S}tanza: A python natural language processing toolkit for many human languages}.
\newblock In \emph{Proceedings of the 58th Annual Meeting of the Association for Computational Linguistics: System Demonstrations}, pages 101--108, Online. Association for Computational Linguistics.

\bibitem[{Qwen et~al.(2025)Qwen, :, Yang, Yang, Zhang, Hui, Zheng, Yu, Li, Liu, Huang, Wei, Lin, Yang, Tu, Zhang, Yang, Yang, Zhou, Lin, Dang, Lu, Bao, Yang, Yu, Li, Xue, Zhang, Zhu, Men, Lin, Li, Tang, Xia, Ren, Ren, Fan, Su, Zhang, Wan, Liu, Cui, Zhang, and Qiu}]{qwen2025qwen25technicalreport}
Qwen, :, An~Yang, Baosong Yang, Beichen Zhang, Binyuan Hui, Bo~Zheng, Bowen Yu, Chengyuan Li, Dayiheng Liu, Fei Huang, Haoran Wei, Huan Lin, Jian Yang, Jianhong Tu, Jianwei Zhang, Jianxin Yang, Jiaxi Yang, Jingren Zhou, and 25 others. 2025.
\newblock \href {https://arxiv.org/abs/2412.15115} {Qwen2.5 technical report}.
\newblock \emph{Preprint}, arXiv:2412.15115.

\bibitem[{Ramm et~al.(2017)Ramm, Lo{\'a}iciga, Friedrich, and Fraser}]{ramm-etal-2017-annotating}
Anita Ramm, Sharid Lo{\'a}iciga, Annemarie Friedrich, and Alexander Fraser. 2017.
\newblock \href {https://aclanthology.org/P17-4001/} {Annotating tense, mood and voice for {E}nglish, {F}rench and {G}erman}.
\newblock In \emph{Proceedings of {ACL} 2017, System Demonstrations}, pages 1--6, Vancouver, Canada. Association for Computational Linguistics.

\bibitem[{Rimsky et~al.(2024)Rimsky, Gabrieli, Schulz, Tong, Hubinger, and Turner}]{rimsky-etal-2024-steering}
Nina Rimsky, Nick Gabrieli, Julian Schulz, Meg Tong, Evan Hubinger, and Alexander Turner. 2024.
\newblock \href {https://doi.org/10.18653/v1/2024.acl-long.828} {Steering llama 2 via contrastive activation addition}.
\newblock In \emph{Proceedings of the 62nd Annual Meeting of the Association for Computational Linguistics (Volume 1: Long Papers)}, pages 15504--15522, Bangkok, Thailand. Association for Computational Linguistics.

\bibitem[{Rogers et~al.(2020)Rogers, Kovaleva, and Rumshisky}]{rogers-etal-2020-primer}
Anna Rogers, Olga Kovaleva, and Anna Rumshisky. 2020.
\newblock \href {https://doi.org/10.1162/tacl_a_00349} {A primer in {BERT}ology: What we know about how {BERT} works}.
\newblock \emph{Transactions of the Association for Computational Linguistics}, 8:842--866.

\bibitem[{Srivastava et~al.(2023)Srivastava, Rastogi, Rao, Shoeb, Abid, Fisch, Brown, Santoro, Gupta, Garriga-Alonso, Kluska, Lewkowycz, Agarwal, Power, Ray, Warstadt, Kocurek, Safaya, Tazarv, Xiang, Parrish, Nie, Hussain, Askell, Dsouza, Slone, Rahane, Iyer, Andreassen, Madotto, Santilli, Stuhlm{\"u}ller, Dai, La, Lampinen, Zou, Jiang, Chen, Vuong, Gupta, Gottardi, Norelli, Venkatesh, Gholamidavoodi, Tabassum, Menezes, Kirubarajan, Mullokandov, Sabharwal, Herrick, Efrat, Erdem, Karaka{\c{s}}, Roberts, Loe, Zoph, Bojanowski, {\"O}zyurt, Hedayatnia, Neyshabur, Inden, Stein, Ekmekci, Lin, Howald, Orinion, Diao, Dour, Stinson, Argueta, Ferri, Singh, Rathkopf, Meng, Baral, Wu, Callison-Burch, Waites, Voigt, Manning, Potts, Ramirez, Rivera, Siro, Raffel, Ashcraft, Garbacea, Sileo, Garrette, Hendrycks, Kilman, Roth, Freeman, Khashabi, Levy, Gonz{\'a}lez, Perszyk, Hernandez, Chen, Ippolito, Gilboa, Dohan, Drakard, Jurgens, Datta, Ganguli, Emelin, Kleyko, Yuret, Chen, Tam, Hupkes, Misra, Buzan, Mollo, Yang, Lee, Schrader, Shutova, Cubuk, Segal, Hagerman, Barnes, Donoway, Pavlick, Rodol{\`a}, Lam, Chu, Tang, Erdem, Chang, Chi, Dyer, Jerzak, Kim, Manyasi, Zheltonozhskii, Xia, Siar, Mart{\'\i}nez-Plumed, Happ{\'e}, Chollet, Rong, Mishra, Winata, de~Melo, Kruszewski, Parascandolo, Mariani, Wang, Jaimovitch-Lopez, Betz, Gur-Ari, Galijasevic, Kim, Rashkin, Hajishirzi, Mehta, Bogar, Shevlin, Schuetze, Yakura, Zhang, Wong, Ng, Noble, Jumelet, Geissinger, Kernion, Hilton, Lee, Fisac, Simon, Koppel, Zheng, Zou, Kocon, Thompson, Wingfield, Kaplan, Radom, Sohl-Dickstein, Phang, Wei, Yosinski, Novikova, Bosscher, Marsh, Kim, Taal, Engel, Alabi, Xu, Song, Tang, Waweru, Burden, Miller, Balis, Batchelder, Berant, Frohberg, Rozen, Hernandez-Orallo, Boudeman, Guerr, Jones, Tenenbaum, Rule, Chua, Kanclerz, Livescu, Krauth, Gopalakrishnan, Ignatyeva, Markert, Dhole, Gimpel, Omondi, Mathewson, Chiafullo, Shkaruta, Shridhar, McDonell, Richardson, Reynolds, Gao, Zhang, Dugan, Qin, Contreras-Ochando, Morency, Moschella, Lam, Noble, Schmidt, He, Oliveros-Col{\'o}n, Metz, Senel, Bosma, Sap, Hoeve, Farooqi, Faruqui, Mazeika, Baturan, Marelli, Maru, Ramirez-Quintana, Tolkiehn, Giulianelli, Lewis, Potthast, Leavitt, Hagen, Schubert, Baitemirova, Arnaud, McElrath, Yee, Cohen, Gu, Ivanitskiy, Starritt, Strube, Sw{\k{e}}drowski, Bevilacqua, Yasunaga, Kale, Cain, Xu, Suzgun, Walker, Tiwari, Bansal, Aminnaseri, Geva, Gheini, T, Peng, Chi, Lee, Krakover, Cameron, Roberts, Doiron, Martinez, Nangia, Deckers, Muennighoff, Keskar, Iyer, Constant, Fiedel, Wen, Zhang, Agha, Elbaghdadi, Levy, Evans, Casares, Doshi, Fung, Liang, Vicol, Alipoormolabashi, Liao, Liang, Chang, Eckersley, Htut, Hwang, Mi{\l}kowski, Patil, Pezeshkpour, Oli, Mei, Lyu, Chen, Banjade, Rudolph, Gabriel, Habacker, Risco, Milli{\`e}re, Garg, Barnes, Saurous, Arakawa, Raymaekers, Frank, Sikand, Novak, Sitelew, Bras, Liu, Jacobs, Zhang, Salakhutdinov, Chi, Lee, Stovall, Teehan, Yang, Singh, Mohammad, Anand, Dillavou, Shleifer, Wiseman, Gruetter, Bowman, Schoenholz, Han, Kwatra, Rous, Ghazarian, Ghosh, Casey, Bischoff, Gehrmann, Schuster, Sadeghi, Hamdan, Zhou, Srivastava, Shi, Singh, Asaadi, Gu, Pachchigar, Toshniwal, Upadhyay, Debnath, Shakeri, Thormeyer, Melzi, Reddy, Makini, Lee, Torene, Hatwar, Dehaene, Divic, Ermon, Biderman, Lin, Prasad, Piantadosi, Shieber, Misherghi, Kiritchenko, Mishra, Linzen, Schuster, Li, Yu, Ali, Hashimoto, Wu, Desbordes, Rothschild, Phan, Wang, Nkinyili, Schick, Kornev, Tunduny, Gerstenberg, Chang, Neeraj, Khot, Shultz, Shaham, Misra, Demberg, Nyamai, Raunak, Ramasesh, vinay~uday prabhu, Padmakumar, Srikumar, Fedus, Saunders, Zhang, Vossen, Ren, Tong, Zhao, Wu, Shen, Yaghoobzadeh, Lakretz, Song, Bahri, Choi, Yang, Hao, Chen, Belinkov, Hou, Hou, Bai, Seid, Zhao, Wang, Wang, Wang, and Wu}]{srivastava2023beyond}
Aarohi Srivastava, Abhinav Rastogi, Abhishek Rao, Abu Awal~Md Shoeb, Abubakar Abid, Adam Fisch, Adam~R. Brown, Adam Santoro, Aditya Gupta, Adri{\`a} Garriga-Alonso, Agnieszka Kluska, Aitor Lewkowycz, Akshat Agarwal, Alethea Power, Alex Ray, Alex Warstadt, Alexander~W. Kocurek, Ali Safaya, Ali Tazarv, and 431 others. 2023.
\newblock \href {https://openreview.net/forum?id=uyTL5Bvosj} {Beyond the imitation game: Quantifying and extrapolating the capabilities of language models}.
\newblock \emph{Transactions on Machine Learning Research}.
\newblock Featured Certification.

\bibitem[{Stickland et~al.(2024)Stickland, Lyzhov, Pfau, Mahdi, and Bowman}]{stickland2024steering}
Asa~Cooper Stickland, Alexander Lyzhov, Jacob Pfau, Salsabila Mahdi, and Samuel~R. Bowman. 2024.
\newblock \href {https://openreview.net/forum?id=tfXIZ8P4ZU} {Steering without side effects: Improving post-deployment control of language models}.
\newblock In \emph{Neurips Safe Generative AI Workshop 2024}.

\bibitem[{Vig and Belinkov(2019)}]{vig-belinkov-2019-analyzing}
Jesse Vig and Yonatan Belinkov. 2019.
\newblock \href {https://doi.org/10.18653/v1/W19-4808} {Analyzing the structure of attention in a transformer language model}.
\newblock In \emph{Proceedings of the 2019 ACL Workshop BlackboxNLP: Analyzing and Interpreting Neural Networks for NLP}, pages 63--76, Florence, Italy. Association for Computational Linguistics.

\bibitem[{Vig et~al.(2020)Vig, Gehrmann, Belinkov, Qian, Nevo, Singer, and Shieber}]{vig2020investigating}
Jesse Vig, Sebastian Gehrmann, Yonatan Belinkov, Sharon Qian, Daniel Nevo, Yaron Singer, and Stuart Shieber. 2020.
\newblock \href {https://proceedings.neurips.cc/paper_files/paper/2020/file/92650b2e92217715fe312e6fa7b90d82-Paper.pdf} {Investigating gender bias in language models using causal mediation analysis}.
\newblock In \emph{Advances in Neural Information Processing Systems}, volume~33, pages 12388--12401. Curran Associates, Inc.

\bibitem[{Waldis et~al.(2024)Waldis, Perlitz, Choshen, Hou, and Gurevych}]{waldis2024holmes}
Andreas Waldis, Yotam Perlitz, Leshem Choshen, Yufang Hou, and Iryna Gurevych. 2024.
\newblock \href {https://doi.org/10.1162/tacl_a_00718} {Holmes: A benchmark to assess the linguistic competence of language models}.
\newblock \emph{Transactions of the Association for Computational Linguistics}, 12:1616--1647.

\bibitem[{Wang et~al.(2023)Wang, Variengien, Conmy, Shlegeris, and Steinhardt}]{wang2023interpretability}
Kevin~Ro Wang, Alexandre Variengien, Arthur Conmy, Buck Shlegeris, and Jacob Steinhardt. 2023.
\newblock \href {https://openreview.net/forum?id=NpsVSN6o4ul} {Interpretability in the wild: a circuit for indirect object identification in {GPT}-2 small}.
\newblock In \emph{The Eleventh International Conference on Learning Representations}.

\bibitem[{Warstadt et~al.(2020)Warstadt, Parrish, Liu, Mohananey, Peng, Wang, and Bowman}]{warstadt2020blimp}
Alex Warstadt, Alicia Parrish, Haokun Liu, Anhad Mohananey, Wei Peng, Sheng-Fu Wang, and Samuel~R. Bowman. 2020.
\newblock \href {https://doi.org/10.1162/tacl_a_00321} {{BL}i{MP}: The benchmark of linguistic minimal pairs for {E}nglish}.
\newblock \emph{Transactions of the Association for Computational Linguistics}, 8:377--392.

\bibitem[{Wu et~al.(2025)Wu, Arora, Geiger, Wang, Huang, Jurafsky, Manning, and Potts}]{wu2025axbenchsteeringllmssimple}
Zhengxuan Wu, Aryaman Arora, Atticus Geiger, Zheng Wang, Jing Huang, Dan Jurafsky, Christopher~D. Manning, and Christopher Potts. 2025.
\newblock \href {https://arxiv.org/abs/2501.17148} {Axbench: Steering llms? even simple baselines outperform sparse autoencoders}.
\newblock \emph{Preprint}, arXiv:2501.17148.

\bibitem[{Zhang et~al.(2025)Zhang, Yu, Zang, Eickhoff, and Pavlick}]{zhang2025the}
Ruochen Zhang, Qinan Yu, Matianyu Zang, Carsten Eickhoff, and Ellie Pavlick. 2025.
\newblock \href {https://openreview.net/forum?id=NCrFA7dq8T} {The same but different: Structural similarities and differences in multilingual language modeling}.
\newblock In \emph{The Thirteenth International Conference on Learning Representations}.

\bibitem[{Zhang et~al.(2020)Zhang, Kishore, Wu, Weinberger, and Artzi}]{zhang2020bertscore}
Tianyi Zhang, Varsha Kishore, Felix Wu, Kilian~Q. Weinberger, and Yoav Artzi. 2020.
\newblock \href {https://openreview.net/forum?id=SkeHuCVFDr} {Bertscore: Evaluating text generation with bert}.
\newblock In \emph{International Conference on Learning Representations}.

\end{thebibliography}

\appendix

\section{Implementation Details}
\subsection{Infrastructure}
The experiments involved the 8 billion parameter model Llama-3.1-8B-Instruct and the 7 billion parameter model Qwen-2.5-7B-instruct. They were run on a single server with
8 NVIDIA RTX A6000 48 GB GPUs with CUDA
Version 12.4 and an AMD EPYC 7413 24-Core
Processor. The total runtime for training the probes and performing the grid search for steering was less than two weeks.

\subsection{Libraries}
See Table~\ref{tab:libraries}.

\begin{table*}[h]
\small
    \centering
    \begin{tabular}{llll}
        \textbf{Usage} & \textbf{Library}  & \textbf{Model} & \textbf{Reference} \\
        \midrule
        Training Linear Probes & \texttt{scikit-learn} && \citet{pedregosa2011scikit}\\
        Linear Discriminant Analysis & \texttt{scikit-learn} && \citet{pedregosa2011scikit}\\
        POS-Tagging & \texttt{stanza} && \citet{qi-etal-2020-stanza} \\
        Propbank Annotations & \texttt{nltk} && \citet{elhadad-2010-book} \\
        & \texttt{spacy} & \textit{en\_core\_web\_lg} & \citet{honnibal2020spacy} \\
        BERTScore & \texttt{bert\_score} & \textit{microsoft/deberta-xlarge-mnli} & \citet{zhang2020bertscore} \\
    \end{tabular}
    \caption{Libraries used for experiments.}
    \label{tab:libraries}
\end{table*}

\section{Related Work on Grammatical Competence of LMs}
\label{app:related_work}
\paragraph{Behavioral Evaluations of Grammatical Knowledge.}
Various benchmarks assess the linguistic knowledge of LLMs. BLiMP \citep{warstadt2020blimp} and MultiBLiMP \citep{jumelet2025multiblimp} evaluate syntactic acceptability via paired sentence probabilities, while HOLMES \citep{waldis2024holmes} consolidates linguistic probing datasets across a range of syntactic phenomena, including tense classification tasks from \citet{conneau2018cram} and \citet{klafka2020spying}. Additional studies use probing methods to analyze specific grammatical categories, such as structural syntax \citep{hewitt-manning-2019-structural, diego2024polar} and aspect in morphologically rich languages \citep{katinskaia-yangarber-2024-probing}.  However, these studies typically model tense as a binary past–present distinction and do not address the full range of categorical tense and aspect distinctions.

\paragraph{Causal and Representational Analyses of Syntax in LMs.}
A complementary line of work investigates how grammatical information is encoded within LMs and how it can be manipulated. Studies have examined internal representations of syntax through attention patterns \citep{vig-belinkov-2019-analyzing} and circuit-level structures \citep{wang2023interpretability, ferrando-costa-jussa-2024-similarity}, revealing how syntactic features are distributed across components of the model. Building on this, causal intervention methods have been used to identify which internal features are functionally relevant to grammatical behavior. CausalGym \citep{arora2024causalgym} tests whether linear representations influence syntactic decisions, while other work targets tense specifically, manipulating feed-forward layers \citep{merullo2024language}, attention heads \citep{zhang2025the}, or sparse autoencoder features \citep{brinkmann2025large} to steer generation. These studies, like the behavioral evaluations, focus primarily on binary tense distinctions and are typically limited to single-token evaluations—with the exception of \citet{brinkmann2025large}, who consider open-ended generation. Prior work has not jointly analyzed tense and aspect within a unified framework. Our work advances this area by studying categorical features in multi-token generation, combining probing, representation space analysis, and causal steering to examine how these concepts are encoded and can be controlled.

\section{Tense and Aspect Overview}
We provide an overview of all possible tense-aspect combinations in the English language in Table~\ref{tab:tense_aspect_combi}.

\label{app:tense_aspect_overview}
\begin{table*}[]
    \centering
    \small
    \begin{tabular}{l|lll}
         & \textbf{present} & \textbf{past} & \textbf{future} \\
             \midrule
         \textbf{simple} & She \textit{drives} her car. & She \textit{drove} her car. & She\textit{ will drive} her car. \\
         \textbf{progressive} & She \textit{is driving} her car. & She \textit{was driving} her car. & She \textit{will be driving} her car. \\
         \textbf{perfect} & She \textit{has driven} her car. & She \textit{had driven} her car. & She \textit{will have driven} her car. \\
         \textbf{perfect progressive} & She \textit{has been driving} her car. & She \textit{had been driving} her car. & She \textit{will have been driving} her car. \\
    \end{tabular}
    \caption{Example sentence conjugated across different tense-aspect combinations.}
    \label{tab:tense_aspect_combi}
\end{table*}

\section{Dataset Composition}
\label{app:dataset}

To ensure high-quality annotations for both tense and aspect, we prioritized careful dataset selection and manual validation. While existing resources such as TimeML \citep{pustejovsky2006timebank} and Universal Dependencies offer valuable linguistic annotations, they were not an ideal fit for our specific needs, either due to differences in annotation focus or incomplete coverage of tense-aspect information. We also explored the TMV annotation tool \citep{ramm-etal-2017-annotating}, but it relies on legacy part-of-speech (POS) and dependency parsing tools, which rendered it incompatible with our pipeline. 
Given these considerations, we curated a dataset with 692 sentences from PropBank \citep{palmer-etal-2005-proposition}, each containing exactly one verb. To enforce this constraint, we applied dependency parsing and filtered out any sentence containing more than one word tagged with the POS label `VERB`. While the filtering is not perfectly precise, it provides a reasonably effective heuristic for isolating single-verb constructions.
To improve the overall data coverage for underrepresented tense-aspect combinations, we generated additional synthetic examples using ChatGPT-4o \citep{hurst2024gpt}, prompted with templates shown in the text box below. After downsampling the train set to address any remaining class imbalance, we have 348 samples per class for tense and 261 samples per class for aspect.

The test set from BIG-bench \citep{srivastava2023beyond, logeswaran2018content} is balanced with $\sim$70 samples per unique aspect and $\sim$90 samples per unique tense.

\begin{tcolorbox}[
    colback=white,
    colframe=black,
    coltitle=white,
    fonttitle=\bfseries,
    sharp corners=all,
    boxrule=0.8pt,
    arc=3mm,
    width=\linewidth,
    title=Prompts for Synthetic Data Generation
]

\begin{tcolorbox}[
    colback=gray!20,
    colframe=gray!20,
    sharp corners=all,
    boxrule=0pt,
    arc=0mm,
    left=1mm,
    right=1mm,
    top=1mm,
    bottom=1mm
]
Generate 100 diverse sentences in the [\textsc{past}] tense. Each sentence should contain only one verb and should vary in structure, subject, and length. 
\end{tcolorbox}
\begin{tcolorbox}[
    colback=gray!20,
    colframe=gray!20,
    sharp corners=all,
    boxrule=0pt,
    arc=0mm,
    left=1mm,
    right=1mm,
    top=1mm,
    bottom=1mm
]
Generate 100 diverse sentences in [\textsc{past}] tense. Each sentence should not contain more than one verb and should vary in structure, subject, and length. 
\end{tcolorbox}
\begin{tcolorbox}[
    colback=gray!20,
    colframe=gray!20,
    sharp corners=all,
    boxrule=0pt,
    arc=0mm,
    left=1mm,
    right=1mm,
    top=1mm,
    bottom=1mm
]
Generate a list of 100 random sentences that are in active or passive voice, declarative or interrogative, singular or plural. Each sentence should contain only one single verb phrase of the tense "[\textsc{past}]".
\end{tcolorbox}
\end{tcolorbox}

\section{Additional Probe Results}
\label{app:probing_heatmaps}
Besides length-normalized sum pooling, we explore naive summation, mean pooling, and the final token representation as alternative methods for extracting representations:

\begin{equation}
    h_{\textrm{sum pooling}} = \sum_{i=1}^N h_i,
\end{equation} 
\begin{equation}
    h_{\textrm{mean pooling}} = \frac{1}{N}\sum_{i=1}^N h_i,
\end{equation} 
\begin{equation}
    h_{\textrm{final token}} = h_{-1}.
\end{equation} 

\begin{figure}[H]
    \centering
    \includegraphics[width=1\linewidth]{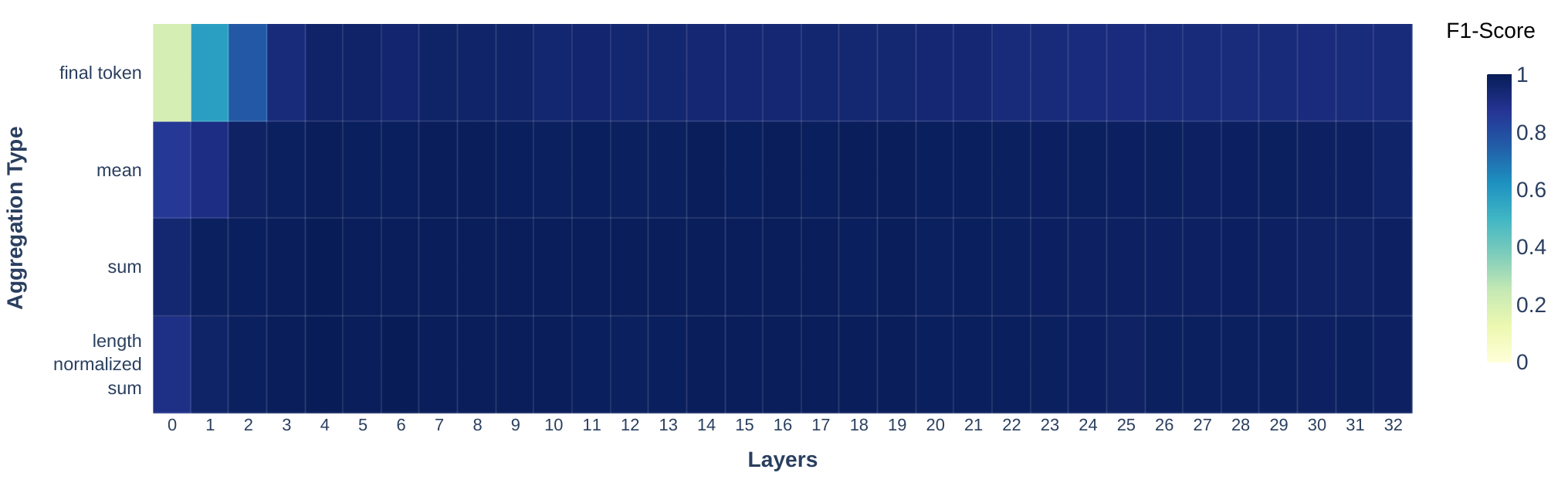}
    \includegraphics[width=1\linewidth]{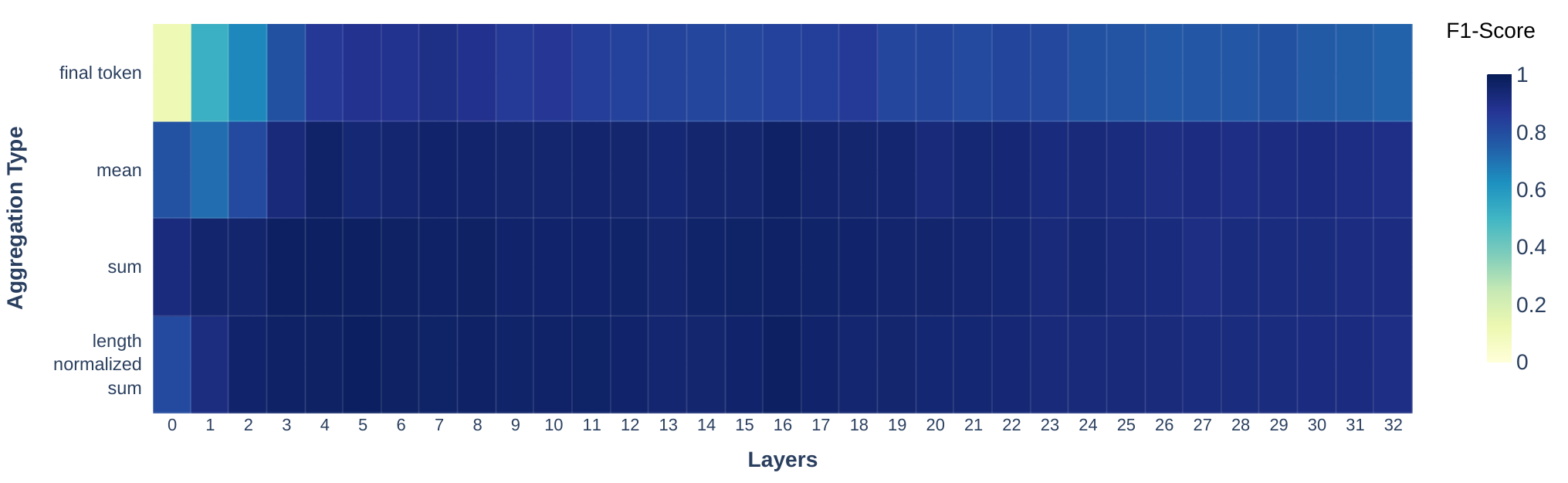}
    \includegraphics[width=1\linewidth]{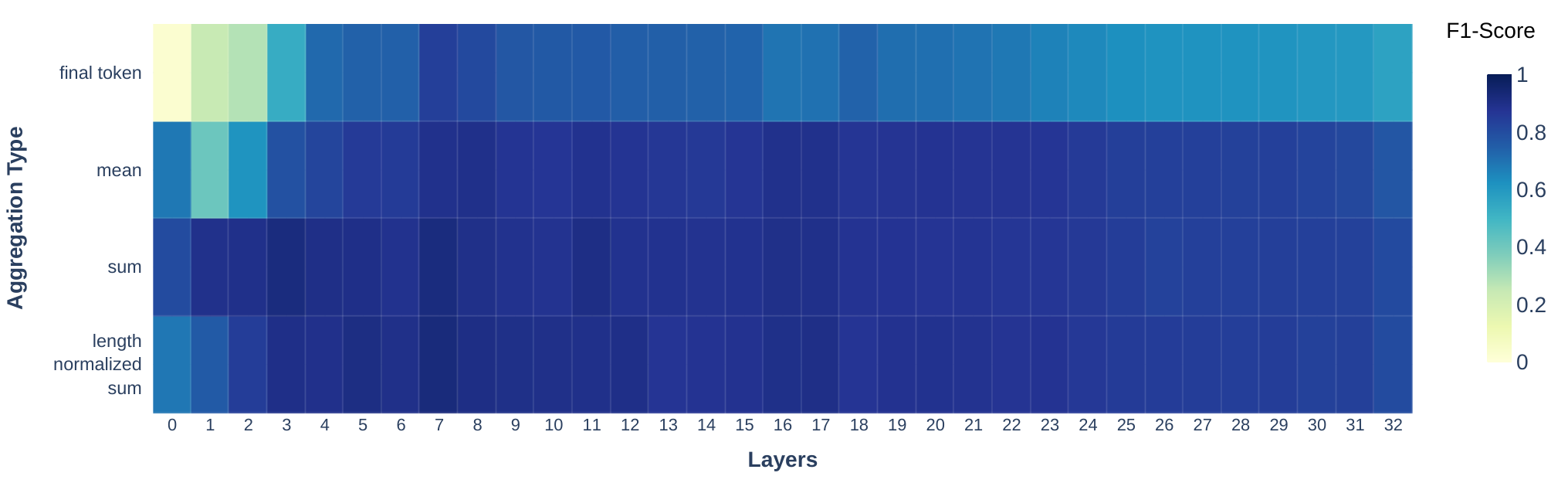}
    \caption{Llama-8B probing f1-scores for \textbf{tense}, \textbf{aspect} and \textbf{tense-aspect} across layers (L0: embedding layer).}
    \label{fig:probing_layerwise_llama}
\end{figure}

\begin{figure}
    \centering
    \includegraphics[width=1\linewidth]{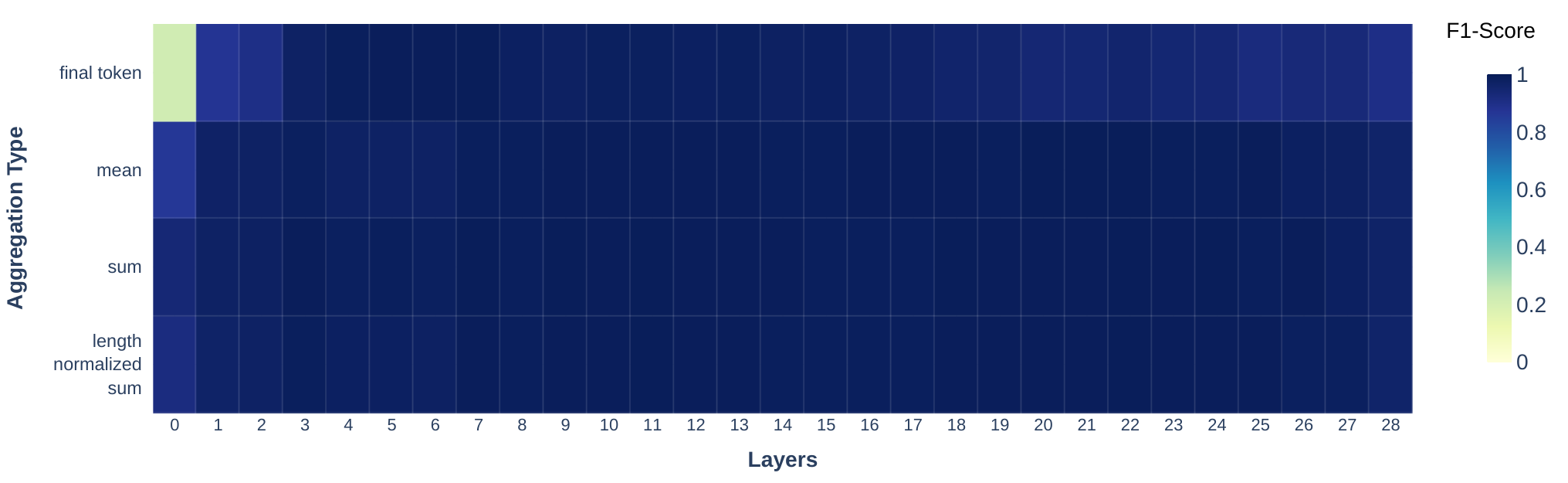}
    \includegraphics[width=1\linewidth]{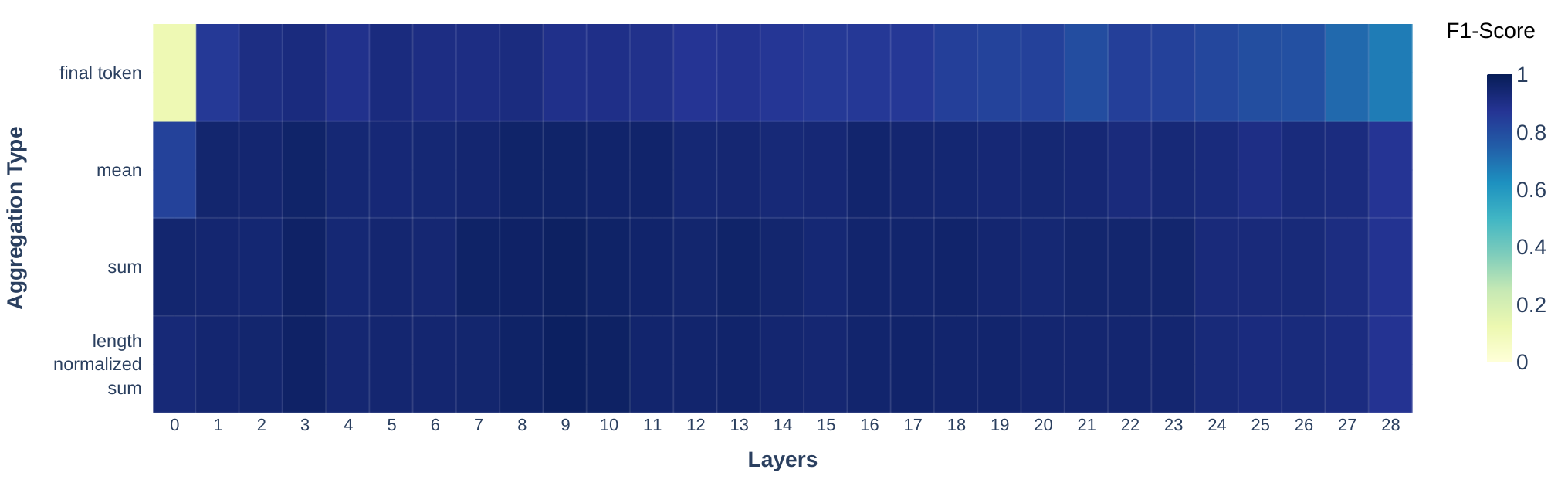}
        \includegraphics[width=1\linewidth]{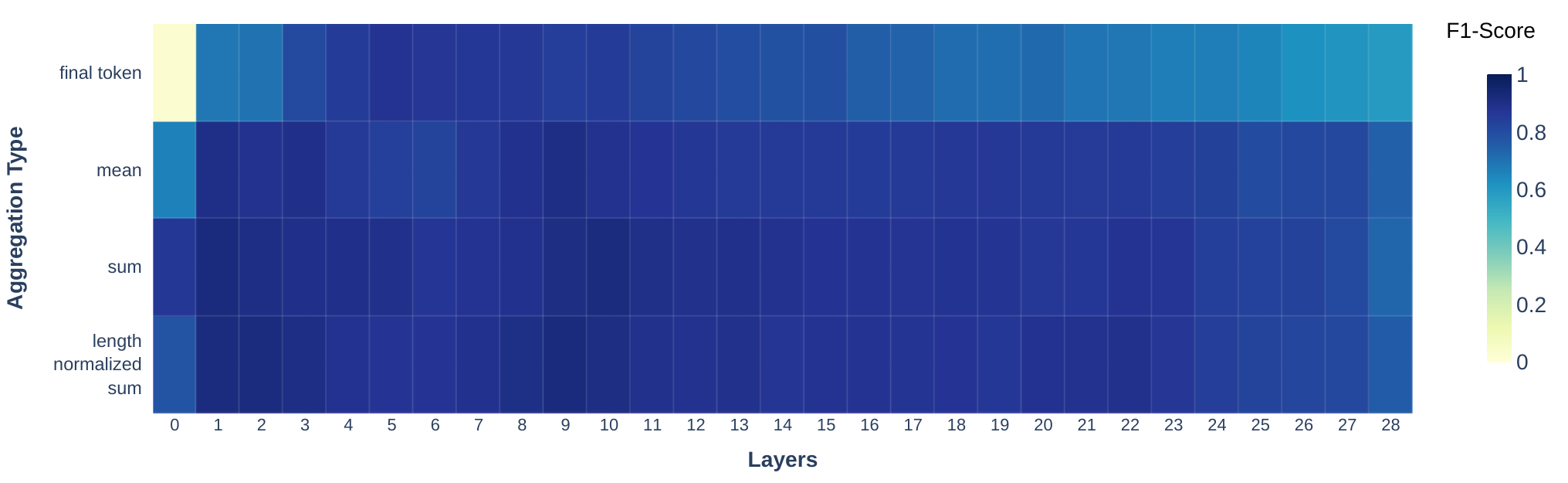}
    \caption{Qwen-7B probing f1-scores for \textbf{tense}, \textbf{aspect} and \textbf{tense-aspect} across layers.}
    \label{fig:probing_layerwise_qwen}
\end{figure}

We provide layer-wise probing results across different aggregation strategies in Figure~\ref{fig:probing_layerwise_llama} for Llama-8B and in Figure~\ref{fig:probing_layerwise_qwen} for Qwen-7B. Across both, models and targets, sum pooling and its length normalized version yield the highest f1-scores. Using the final token representation only is slightly less informative and limited to the earlier middle layers.

\section{Cluster Quality of LDA Projections}
\label{app:cluster_quality}
To quantify the separation effectiveness of our LDA directions, we compute the following metrics for the L0 projected hidden states and report them in Table~\ref{tab:cluster_quality_metrics} (visualizations of Qwen-7B: Fig.~\ref{fig:lda_tense_dirs},~\ref{fig:lda_aspect_dirs}):

\begin{itemize}
    \item \textbf{Explained Variance ($\uparrow$):} ratio of between-class variance and total variance
    \item \textbf{Fisher Discriminant Ratio ($\uparrow$):} between-class vs within-class variance
    \item \textbf{Silhouette score ($\uparrow$):} similarity of a point to its own class vs others.
\end{itemize}

\begin{table*}[h]
    \centering
    \begin{tabular}{lllll}
        & \multicolumn{2}{c}{\textsc{Qwen-7B}} & \multicolumn{2}{c}{\textsc{Llama-8B}} \\
        \midrule
         & \textbf{Tense} & \textbf{Aspect} & \textbf{Tense} & \textbf{Aspect}\\
         \midrule
         Explained Variance & 0.72 & 0.70 & 0.63 & 0.62\\
         Fisher Discriminant Ratio & 2.44 & 2.46 & 1.53 & 1.78\\
         Silhouette score & 0.39 & 0.24& 0.27 & 0.22 \\
    \end{tabular}
    \caption{Cluster quality scores for the L0 projected hidden states indicate that at least 70\% of total variance is explained through the retrieved tense and aspect classes for Qwen-7B and at least 62\% for Llama-8B. The Fisher ratios demonstrate that between-class variance exceeds within-class variance by 1.5 - 2.5, confirming that our LDA directions successfully capture the target linguistic distinctions. While these scores indicate moderate rather than perfect separation, this is reasonable since tense and aspect categories can have fuzzy boundaries (e.g., “Tomorrow I leave for Paris.” uses present form but future reference).}
    \label{tab:cluster_quality_metrics}
\end{table*}

Further, to measure the orthogonality of the parent feature directions $\bar{\ell}_{\text{\scriptsize\textsc{tense}}}$ and $\bar{\ell}_{\text{\scriptsize\textsc{aspect}}}$, we compute the cosine similarity between all possible vector differences for tense and aspect respectively, considering the layers with the best steering efficacy across tasks and methods. This results in a mean similarity of 0.045 for Llama-8B and 0.118 for Qwen-7B. Cosine values this small imply that the two contrast directions are almost orthogonal, so the models largely encode tense and aspect in independent subspaces.

\section{Tasks for Multi-Token Steering}
\label{app:testset_steering}
We consider three generative tasks to evaluate the steering procedure, their prompt formats are detailed below. 

\begin{tcolorbox}[
    colback=white, 
    colframe=black, 
    coltitle=white,
    fonttitle=\bfseries,
    sharp corners=all,
    boxrule=0.8pt,
    arc=3mm,
    width=\linewidth,
    title=Random Sentence Task
]
Using the template "<\texttt{Imperative Verb}> <\texttt{Sentence Description}>:" with values from below, we form a test set of $N=83$ distinct prompts for the random sentence task (e.g., "Formulate one grammatically correct sentence:"). 

\begin{tcolorbox}[
    colback=gray!20,
    colframe=gray!20,
    sharp corners=all,
    boxrule=0pt,
    arc=0mm,
    left=1mm,
    right=1mm,
    top=1mm,
    bottom=1mm
]
\textbf{Imperative Verbs}\\
Generate, Create, Produce, Write, Output, Provide, Construct, Make up, Formulate, Come up, Print, Return, Craft
\end{tcolorbox}

\begin{tcolorbox}[
    colback=gray!20,
    colframe=gray!20,
    sharp corners=all,
    boxrule=0pt,
    arc=0mm,
    left=1mm,
    right=1mm,
    top=1mm,
    bottom=1mm
]
\textbf{Sentence Descriptions}\\
a single sentence, one sentence, a random sentence, a sentence using any verb tense, an arbitrary sentence, one grammatically correct sentence
\end{tcolorbox}
\end{tcolorbox}

For repetition and temporal translation, we only include samples where the unsteered output is a valid answer.

\begin{tcolorbox}[
    colback=white, 
    colframe=black, 
    coltitle=white,
    fonttitle=\bfseries,
    sharp corners=all,
    boxrule=0.8pt,
    arc=3mm,
    width=\linewidth,
    title=Few-Shot Tasks
]
We create a prompt for each sentence in our test set and use other sentences from the test set as few-shot examples. For each steering target, we exclude those samples, where the source feature value is equal to the steering target, resulting in a test set size of $N=211$ for aspect and $N=191$ for tense.

\begin{tcolorbox}[
    colback=gray!20,
    colframe=gray!20,
    sharp corners=all,
    boxrule=0pt,
    arc=0mm,
    left=1mm,
    right=1mm,
    top=1mm,
    bottom=1mm
]
\textbf{Repetition Task}\\
I am writing a story. \textbackslash \textbackslash{} I am writing a story. \\\\ I have finished. \textbackslash \textbackslash{} I have finished. \\\\ The dog is barking. \textbackslash \textbackslash{}
\end{tcolorbox}

\begin{tcolorbox}[
    colback=gray!20,
    colframe=gray!20,
    sharp corners=all,
    boxrule=0pt,
    arc=0mm,
    left=1mm,
    right=1mm,
    top=1mm,
    bottom=1mm
]
\textbf{Temporal translation Task}\\
I have been walking through the park. \textbackslash \textbackslash{} I have walked through the park. \\\\ Paul has been visiting the school. \textbackslash \textbackslash{} Paul has visited the school. \\\\ He has been earning a six figure salary. \textbackslash \textbackslash{} 
\end{tcolorbox}
\end{tcolorbox}

\section{Grid search over Steering Factor}
\label{app:gridsearch_steering}
We use the random sentence task to perform the initial grid search across the $\alpha$ values listed in Table~\ref{tab:alpha_grid}, and use the findings to adjust the search space for the few-shot tasks accordingly. The best steering configurations for each model, task, method and target are listed in Table~\ref{tab:optimal_steering}.

\begin{table*}[]
    \centering
     \begin{tabular}{llp{8.5cm}}
         \textbf{Task} & \textbf{Model} & \textbf{$\alpha$ values} \\
         \midrule
        Random Sentence & both & 0.1, 0.5, 1, 1.5, 2, 3, 4, 5, 7, 10, 15, 20, 25, 30, 50 \\
        Random Sentence & Qwen-2-7B-Instruct & 100, 150, 200, 250, 300, 400, 500\\
        Few-Shot & Llama-3-8B-Instruct & 5, 7, 10, 15, 20, 25, 30, 35, 40 \\
        Few-Shot & Qwen-2-7B-Instruct & 200, 225, 250, 275, 300, 325, 350, 375, 400, 425, 450, 475, 500, 550, 700, 800 \\
    \end{tabular}
    \caption{$\alpha$ values searched across in grid search for steering experiments.}
    \label{tab:alpha_grid}
\end{table*}

\begin{table*}[]
    \centering
    \begin{tabular}{llllll}
        \textbf{Model} & \textbf{Task} & \textbf{Method} & \textbf{Target} & \textbf{Layer} & \textbf{Alpha} \\
        \midrule
        Llama-3-8B-Instruct & Random & TA & Tense & 13 & 5 \\
        Llama-3-8B-Instruct & Repetition & TA & Tense & 11 & 15 \\
        Llama-3-8B-Instruct & Repetition & TA+SS & Tense & 12 & 10 \\
        Llama-3-8B-Instruct & Repetition & TA+Proj-SS & Tense & 11 & 15 \\
        Llama-3-8B-Instruct & Temporal Translation & TA & Tense & 13 & 5 \\
        Llama-3-8B-Instruct & Temporal Translation & TA+SS & Tense & 12 & 7 \\
        Llama-3-8B-Instruct & Temporal Translation & TA & Tense & 12 & 10 \\
        Llama-3-8B-Instruct & Random & TA & Aspect & 19 & 15 \\
        Llama-3-8B-Instruct & Repetition & TA & Aspect & 14 & 15 \\
        Llama-3-8B-Instruct & Repetition & TA+SS & Aspect & 19 & 15 \\
        Llama-3-8B-Instruct & Repetition & TA+Proj-SS & Aspect & 14 & 15 \\
        Llama-3-8B-Instruct & Temporal Translation & TA & Aspect & 18 & 20 \\
        Llama-3-8B-Instruct & Temporal Translation & TA+SS & Aspect & 16 & 15 \\
        Llama-3-8B-Instruct & Temporal Translation & TA & Aspect & 18 & 25 \\
        \midrule
        Qwen-2-7B-Instruct & Random & TA & Tense & 20 & 100 \\
        Qwen-2-7B-Instruct & Repetition & TA & Tense & 22 & 200 \\
        Qwen-2-7B-Instruct & Repetition & TA+SS & Tense & 24 & 200 \\
        Qwen-2-7B-Instruct & Repetition & TA+Proj-SS & Tense & 22 & 225 \\
        Qwen-2-7B-Instruct & Temporal Translation & TA & Tense & 22 & 200 \\
        Qwen-2-7B-Instruct & Temporal Translation & TA+SS & Tense & 24 & 200 \\
        Qwen-2-7B-Instruct & Temporal Translation & TA & Tense & 22 & 200 \\
        Qwen-2-7B-Instruct & Random & TA & Aspect & 21 & 150 \\
        Qwen-2-7B-Instruct & Repetition & TA & Aspect & 21 & 200 \\
        Qwen-2-7B-Instruct & Repetition & TA+SS & Aspect & 23 & 200 \\
        Qwen-2-7B-Instruct & Repetition & TA+Proj-SS & Aspect & 21 & 200 \\
        Qwen-2-7B-Instruct & Temporal Translation & TA & Aspect & 22 & 225 \\
        Qwen-2-7B-Instruct & Temporal Translation & TA+SS & Aspect & 24 & 200 \\
        Qwen-2-7B-Instruct & Temporal Translation & TA & Aspect & 22 & 225 \\
        
    \end{tabular}
    \caption{Best steering configuration with regard to efficacy.}
    \label{tab:optimal_steering}
\end{table*}

\section{Measuring Degenerates}
\label{app:steering_metrics}
To measure the rate of degenerate outputs during steering experiments, we track different n-gram statistics. We label an output as degenerate, if it does not pass all the filters in Table~\ref{tab:ngrams} and/or does not contain a verb phrase (i.e., AUX / VERB), as detected by stanza's POS-tagger. N-gram diversity is computed as the product of one minus the repetition rates of 2-, 3-, and 4-grams in the text. These metrics to measure text diversity are based on \citet{li-etal-2023-contrastive}.

\begin{table}[]
    \centering
    \begin{tabular}{ll}
         \textbf{Filter} & \textbf{Threshold} \\
         \midrule
         Unigram Repetition Rate &  < 0.25 \\
         2-gram Repetition Rate& < 0.3\\
         4-gram Repetition Rate& < 0.2\\
         N-gram Repetition Diversity & > 0.5 \\
    \end{tabular}
    \caption{Thresholds for degeneration-filter.}
    \label{tab:ngrams}
\end{table}

\section{Additional Steering Results}
\label{app:additional_steering_results}
\subsection{Selectivity.}
\label{app:selectivity}
\begin{figure}
    \centering
    \includegraphics[width=1\linewidth]{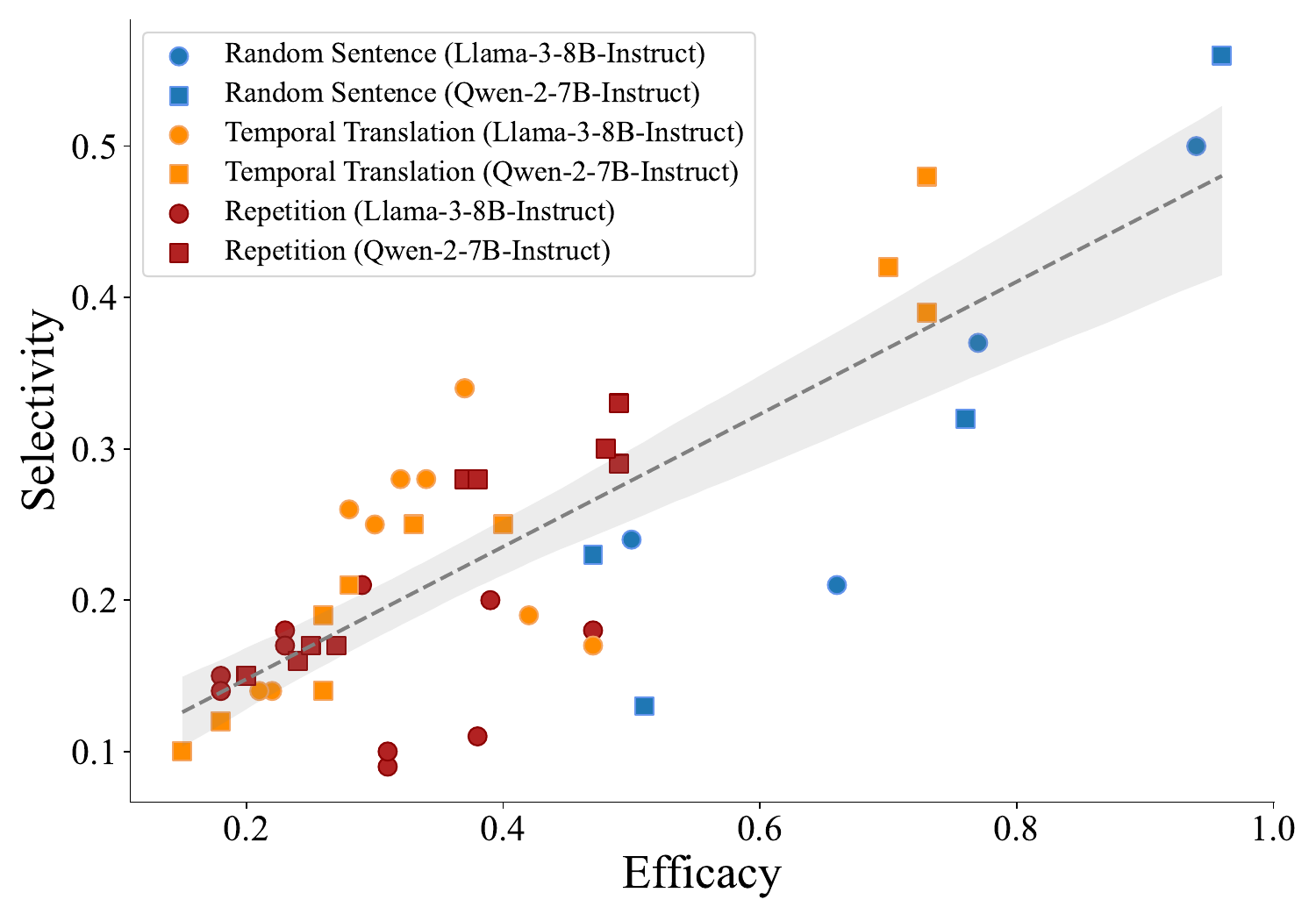}
    \caption{Correlation between efficacy and selectivity is apparent.}
    \label{fig:selectivity}
\end{figure}

The most effective steering methods also show the highest functional selectivity (Figure~\ref{fig:selectivity}), indicating that it is possible to steer one verb property---such as tense---without necessarily affecting the other, like aspect. However, selectivity remains below 50\% on average. Despite the orthogonal representations of tense and aspect, steering one target often influences both. This suggests that the intervention may still be too large, modifying more of the activation than intended.

\subsection{Relative Perplexity Change}
\label{app:perplexity}
\begin{figure}
    \centering
    \includegraphics[width=1\linewidth]{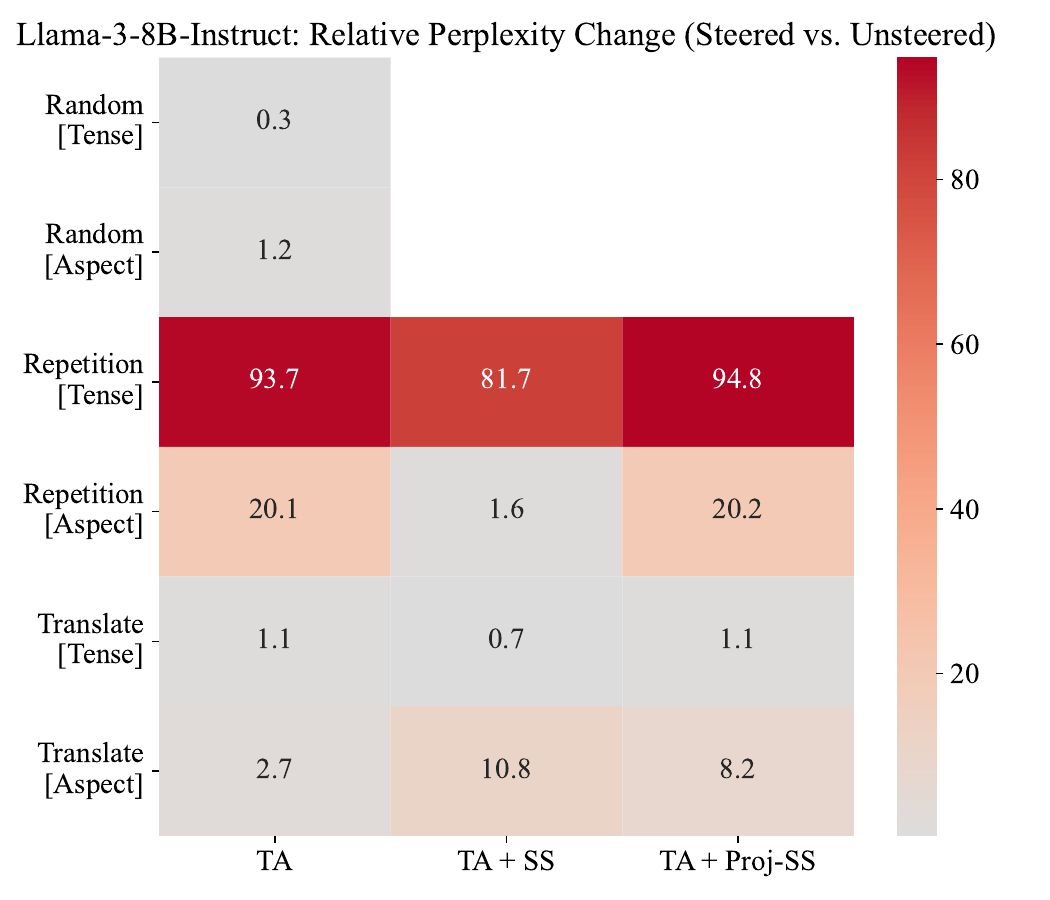}
    \includegraphics[width=1\linewidth]{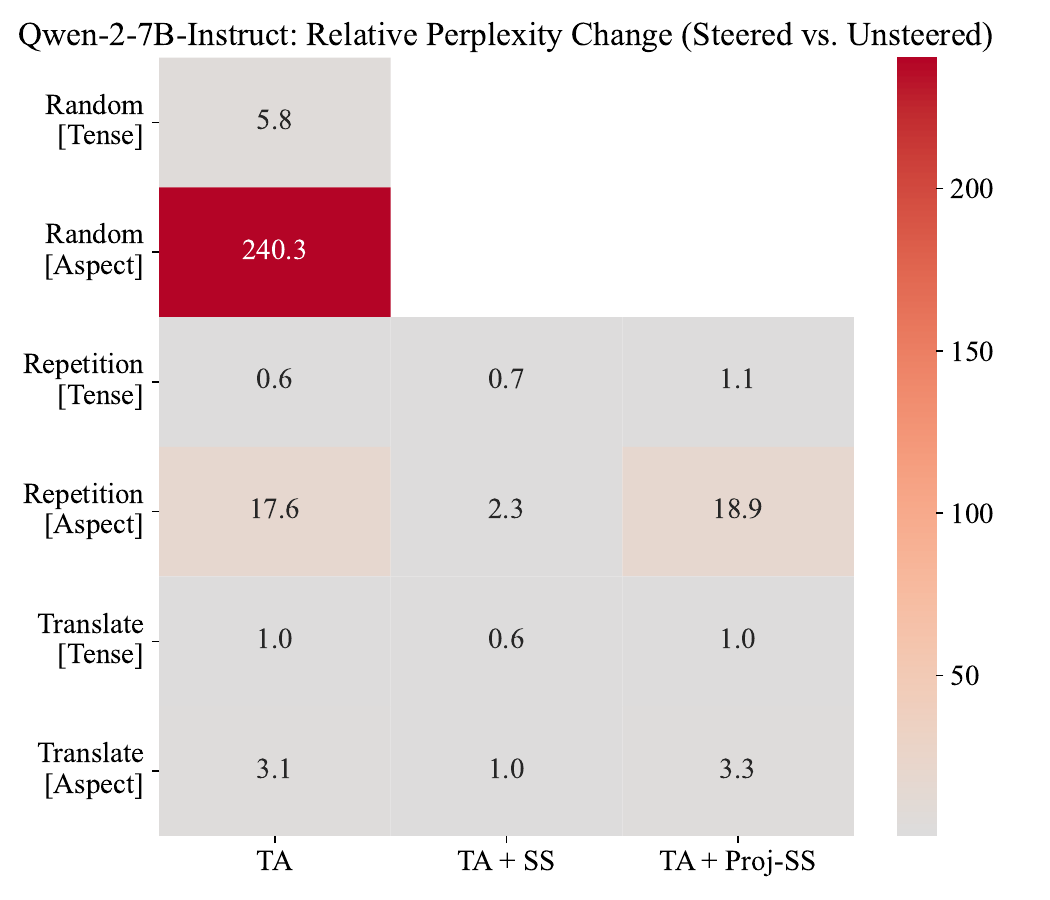}
    \caption{Relative change in perplexity for both models. Overall, the different steering methods lead to minor perplexity increases across tasks, targets and models. However, steering the repetition task appears to be less stable, with outliers across all methods. Similarly, generation quality collapses for steering aspect during random generation of Qwen-7B. This shows that tasks with different internal mechanisms react differently when steered with the same feature vectors and highlights the need for diverse testbeds when developing steering approaches.}
    \label{fig:ppl_change}
\end{figure}

In addition to the degenerate rate which is implicitly measured through efficacy, we report the relative change in perplexity between steered and unsteered generation. Results for the best setups per model, task, target and method are reported in Figure~\ref{fig:ppl_change}. For both models, the majority of setups (8/14 for Llama-8B and 11/14 for Qwen-7B) leads to a minor increase in perplexity with a relative change of < 10. 

\subsection{Activation Norm}
\label{app:activation_norm}
\begin{figure}
    \centering
    \includegraphics[width=1\linewidth]{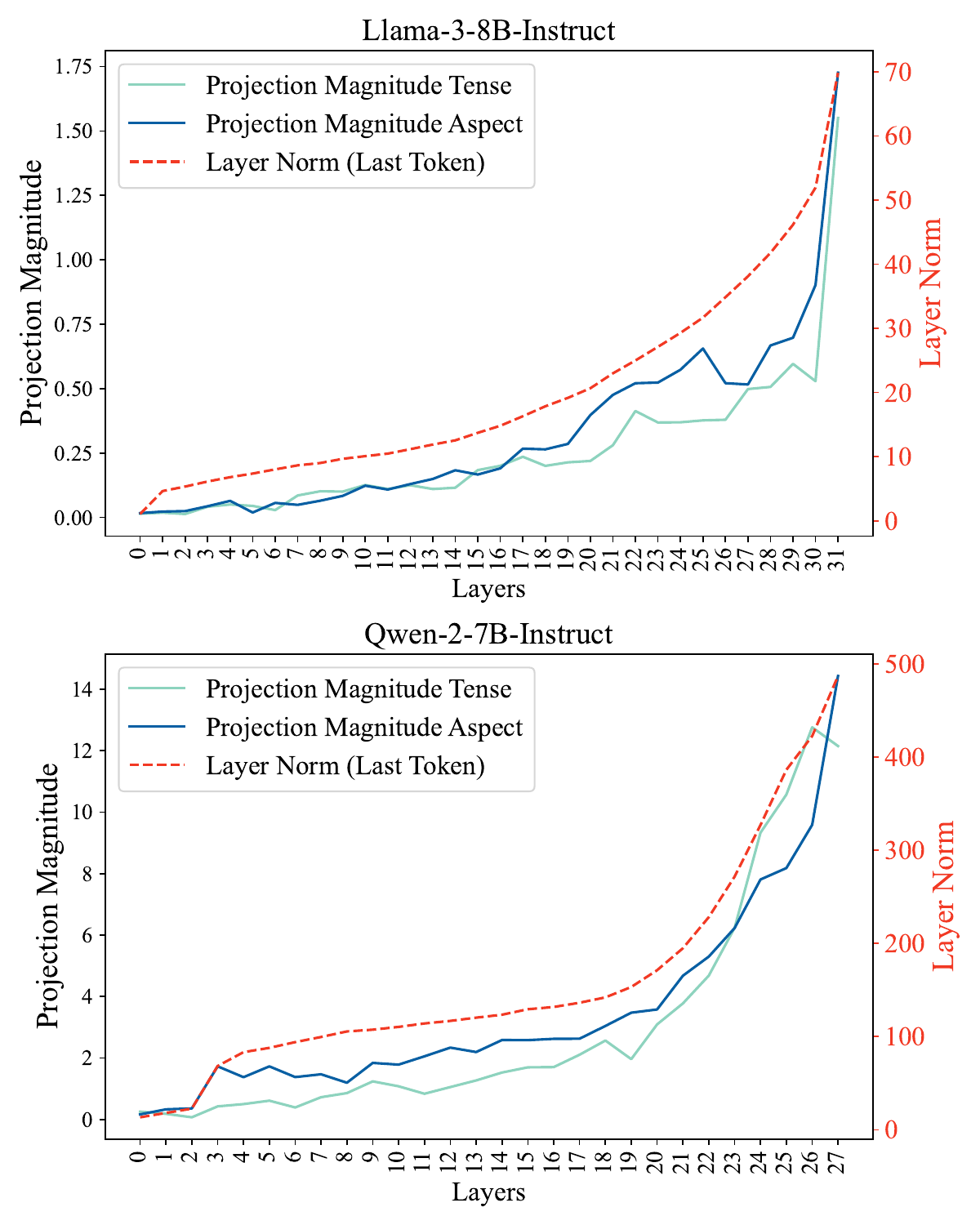}
    \caption{Average activation norm on final token compared to projection magnitude of both features, tense and aspect, averaged across tasks. The graph shows an increase of all three across layers, indicating that the strength of a feature signal roughly scales with the general activation norm.}
    \label{fig:norm_per_task}
\end{figure}

\begin{figure*}
    \centering
    \includegraphics[width=1\linewidth]{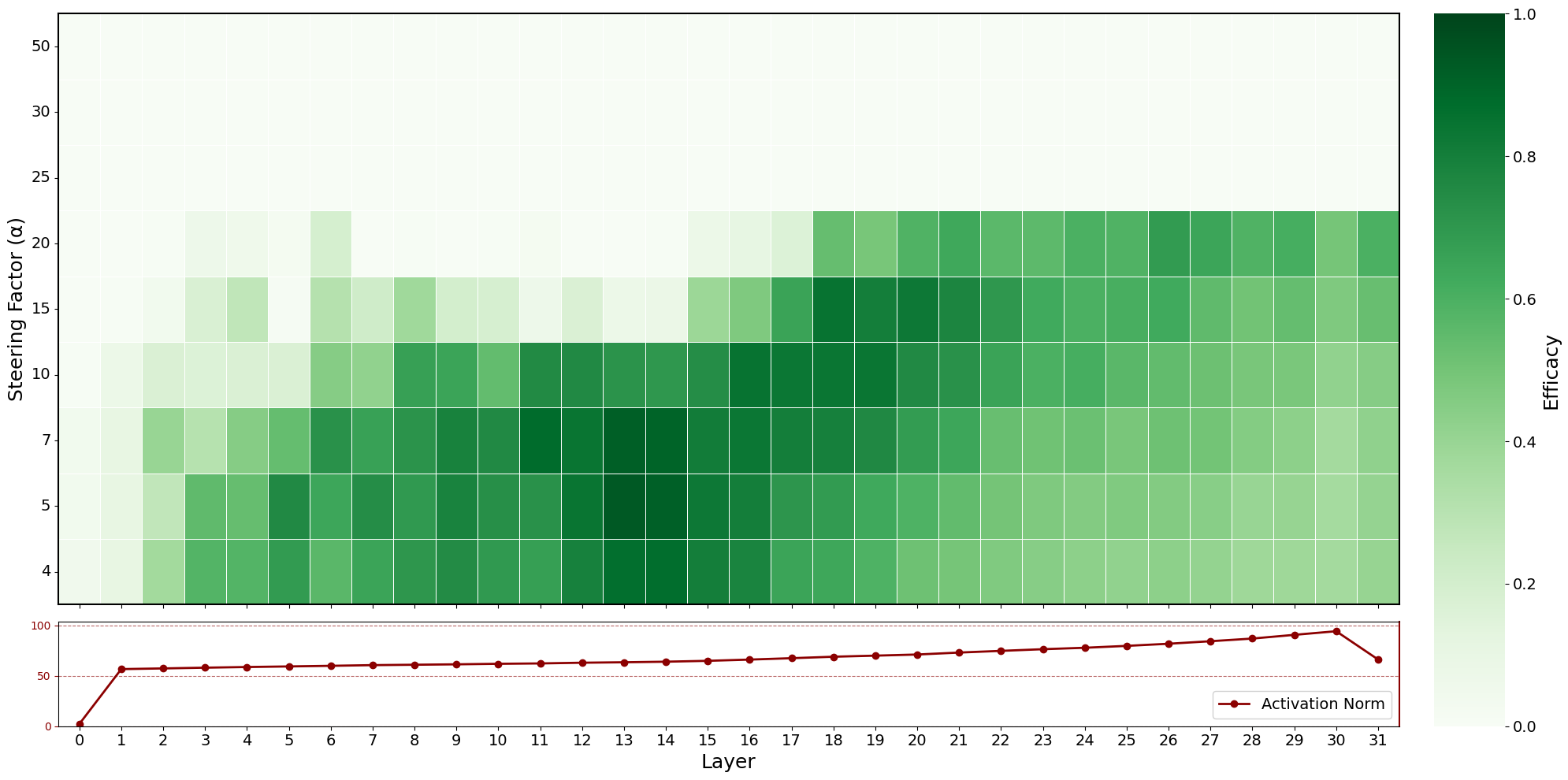}
    \caption{Grid search over best layer and alpha for steering success in Llama-8B. Heatmap displays efficacy with darker colors indicating higher steering success. Bottom plot indicates the average norm of activations per layer.}
    \label{fig:llama_gridsearch}
\end{figure*}

There are differences in activation norms across models as well as across layers of the same model that affect the required steering strength for successful interventions. For an overview of average activation norms and feature projection magnitudes per model, see Figure~\ref{fig:norm_per_task}. The results of a comprehensive grid search across layers and different steering factors is visualized in Figure~\ref{fig:llama_gridsearch}.

\subsection{Nucleus Sampling}
Due to the increased risk of degeneration introduced through steering \citep{stickland2024steering}, we additionally evaluate model behavior under stochastic decoding using nucleus sampling with a temperature of 0.7, but find the effects on efficacy and degenerates to be not consistent, suggesting that the decoding strategy alone cannot prevent degeneration caused by steering.

\subsection{Qualitative Results for Temporal Translation}
\label{app:steering_examples_tt}
Example outputs for temporal translation task are provided in Table~\ref{tab:translation_examples}, demonstrating successful steering of aspect, while keeping tense and the general topic of the sentence fixed. Ablations results for position-wise steering are given in Table~\ref{tab:translation_pos}.

\begin{table*}[]
\small
\begin{tabular}{p{3.1cm}|p{5.4cm}|p{1.6cm}|p{3.5cm}}
\toprule
 \textbf{Temporal Translation} & \textbf{Prompt} & \textbf{Target} & \textbf{Output} \\
 \midrule
\colorbox{purple!25}{Perfect Progressive} $\rightarrow$ \colorbox{orange!25}{Perfect} & 
\begin{tabular}[t]{@{}l@{}}
\colorbox{purple!25}{I have been walking through the park.} \textbackslash\textbackslash{} \\\colorbox{orange!25}{I have walked through the park.} \\\\
\colorbox{purple!25}{Paul has been visiting the school.} \textbackslash\textbackslash{} \\\colorbox{orange!25}{Paul has visited the school.} \\\\
\colorbox{purple!25}{He has been earning a six figure salary.} \textbackslash\textbackslash{} 
\end{tabular}
& \colorbox{cyan!25}{Progressive} & 
\multirow{2}{=}{\parbox{3.5cm}{
\colorbox{orange!25}{\textbf{Unsteered:}} He has earned a six figure salary. \\
\colorbox{cyan!25}{\textbf{Steered:}} He is earning a six figure salary. 
}} \\
\midrule
\colorbox{purple!25}{Perfect} $\rightarrow$ \colorbox{orange!25}{Progressive} & 
\begin{tabular}[t]{@{}l@{}}
\colorbox{purple!25}{I have walked through the park.} \textbackslash\textbackslash{} \\\colorbox{orange!25}{I am walking through the park.} \\\\
\colorbox{purple!25}{Paul has visited the school.} \textbackslash\textbackslash{} \\\colorbox{orange!25}{Paul is visiting the school.} \\\\
\colorbox{purple!25}{He will not have passed the test.} \textbackslash\textbackslash{} 
\end{tabular}
& \colorbox{cyan!25}{Simple} & 
\multirow{2}{=}{\parbox{3.5cm}{
\colorbox{orange!25}{\textbf{Unsteered:}} He will not be passing the test. \\
\colorbox{cyan!25}{\textbf{Steered:}} He will not pass the test. 
}} \\
 \bottomrule
\end{tabular}
\caption{Qualitative examples of steering aspect in Qwen-2-7B-Instruct on the temporal transformation task. Red indicates the aspect of the original sentence, orange the aspect in the unsteered translation and blue marks the aspect that is expected after steering.}
\label{tab:translation_examples}
\end{table*}

 \begin{table*}[]
\small
\setlength{\tabcolsep}{4pt} 
\begin{tabular}{lllll}
\toprule
 \textbf{Steering Position} & \textbf{Prompt Tokens} & \textbf{Generated Tokens} & \textbf{Output (TA)} & \textbf{Output (TA-SS)}\\
 \midrule
 all verb tokens in prompt & ...\texttt{It} \colorbox{orange!25}{\texttt{was snow ing}} \texttt{.} \texttt{\textbackslash\textbackslash} &  \texttt{It} \texttt{is} \texttt{snow} \texttt{ing} \texttt{.} & It is snowing. & It is snowing.\\
 
 last verb token in prompt & ...\texttt{It was snow} \colorbox{orange!25}{\texttt{ing}} \texttt{.} \texttt{\textbackslash\textbackslash} &  \texttt{It} \texttt{is snow ing} \texttt{.} & It is snowing. & It is snowing. \\
 
   sentence end in prompt & ...\texttt{It was snow} \texttt{ing}\colorbox{orange!25}{\texttt{.}} \texttt{\textbackslash\textbackslash} &  \texttt{It is snow ing .}& It is snowing. & It is snowing.\\
   
   final token in prompt & ...\texttt{It was snow ing .} \colorbox{orange!25}{\texttt{\textbackslash\textbackslash}} &  \texttt{It} \texttt{is} \texttt{snow} \texttt{ing} \texttt{.} & It is snowing. & It is snowing.\\
   
final tokens during generation & ...\texttt{It was snow ing .} \colorbox{orange!25}{\texttt{\textbackslash\textbackslash}} &  \colorbox{orange!25}{\texttt{It}} \colorbox{orange!25}{\texttt{is}} \colorbox{orange!25}{\texttt{snow}} \colorbox{orange!25}{\texttt{ing}} \texttt{.} & \textcolor{purple!80}{It was a year...} & \textcolor{purple!30!orange}{It was raining.}\\

  generated token before verb & ...\texttt{It was snow ing .} \texttt{\textbackslash\textbackslash} &  \colorbox{orange!25}{\texttt{It}} \texttt{is} \texttt{snow} \texttt{ing} \texttt{.} & \textcolor{green!60!black}{It was snowing.} & \textcolor{green!60!black}{It was snowing.} \\
  
 first generated verb token & ...\texttt{It was snow ing .} \texttt{\textbackslash\textbackslash} &  \texttt{It} \colorbox{orange!25}{\texttt{is}} \texttt{snow} \texttt{ing} \texttt{.} & It is snowing. & It is snowing. \\
 
 all generated verb token & ...\texttt{It was snow ing .} \texttt{\textbackslash\textbackslash} &  \texttt{It} \colorbox{orange!25}{\texttt{is snow ing}} \texttt{.} & It is snowing. & It is snowing. \\
 \bottomrule
 \end{tabular}
 \caption{Steering Llama-3-8B-Instruct on the temporal translation task towards past tense for the prompt: "He was crying. \textbackslash\textbackslash{} He is crying.\texttt{\textbackslash n}\texttt{\textbackslash n} We were having dinner. \textbackslash\textbackslash{} We are having dinner. \texttt{\textbackslash n}\texttt{\textbackslash n} It was snowing. \textbackslash\textbackslash{}". \textit{Generated Tokens} refers to the unsteered output.}
 \label{tab:translation_pos}
 \end{table*}

\end{document}